\documentclass[10pt,twocolumn,letterpaper]{article}

\usepackage{iccv}
\usepackage{times}
\usepackage{epsfig}
\usepackage{graphicx}
\usepackage{amsmath}
\usepackage{amssymb}

\usepackage{multirow}
\usepackage{booktabs}
\usepackage{bbm}
\usepackage[accsupp]{axessibility}  

\usepackage[pagebackref=true,breaklinks=true,colorlinks,bookmarks=false]{hyperref}

\iccvfinalcopy 

\ificcvfinal\fi

\begin{document}

\title{RICO: Regularizing the Unobservable for Indoor Compositional Reconstruction}

\author{
Zizhang Li$^{1}$,
~Xiaoyang Lyu$^{2}$,
~Yuanyuan Ding$^{1}$,
~Mengmeng Wang$^{1}$,
~Yiyi Liao$^{1}$\thanks{Corresponding authors.},
~Yong Liu$^{1*}$
\vspace{0.8em}\\
$^{1}$ Zhejiang University, $^{2}$ The University of Hong Kong
}

\maketitle
\ificcvfinal\thispagestyle{empty}\fi

\newcommand{\netname}{\textbf{RICO}}

\newcommand{\ba}{\mathbf{a}}\newcommand{\bA}{\mathbf{A}}
\newcommand{\bb}{\mathbf{b}}\newcommand{\bB}{\mathbf{B}}
\newcommand{\bc}{\mathbf{c}}\newcommand{\bC}{\mathbf{C}}
\newcommand{\bd}{\mathbf{d}}\newcommand{\bD}{\mathbf{D}}
\newcommand{\be}{\mathbf{e}}\newcommand{\bE}{\mathbf{E}}
\newcommand{\bff}{\mathbf{f}}\newcommand{\bF}{\mathbf{F}} %
\newcommand{\bg}{\mathbf{g}}\newcommand{\bG}{\mathbf{G}}
\newcommand{\bh}{\mathbf{h}}\newcommand{\bH}{\mathbf{H}}
\newcommand{\bi}{\mathbf{i}}\newcommand{\bI}{\mathbf{I}}
\newcommand{\bj}{\mathbf{j}}\newcommand{\bJ}{\mathbf{J}}
\newcommand{\bk}{\mathbf{k}}\newcommand{\bK}{\mathbf{K}}
\newcommand{\bl}{\mathbf{l}}\newcommand{\bL}{\mathbf{L}}
\newcommand{\bm}{\mathbf{m}}\newcommand{\bM}{\mathbf{M}}
\newcommand{\bn}{\mathbf{n}}\newcommand{\bN}{\mathbf{N}}
\newcommand{\bo}{\mathbf{o}}\newcommand{\bO}{\mathbf{O}}
\newcommand{\bp}{\mathbf{p}}\newcommand{\bP}{\mathbf{P}}
\newcommand{\bq}{\mathbf{q}}\newcommand{\bQ}{\mathbf{Q}}
\newcommand{\br}{\mathbf{r}}\newcommand{\bR}{\mathbf{R}}
\newcommand{\bs}{\mathbf{s}}\newcommand{\bS}{\mathbf{S}}
\newcommand{\bt}{\mathbf{t}}\newcommand{\bT}{\mathbf{T}}
\newcommand{\bu}{\mathbf{u}}\newcommand{\bU}{\mathbf{U}}
\newcommand{\bv}{\mathbf{v}}\newcommand{\bV}{\mathbf{V}}
\newcommand{\bw}{\mathbf{w}}\newcommand{\bW}{\mathbf{W}}
\newcommand{\bx}{\mathbf{x}}\newcommand{\bX}{\mathbf{X}}
\newcommand{\by}{\mathbf{y}}\newcommand{\bY}{\mathbf{Y}}
\newcommand{\bz}{\mathbf{z}}\newcommand{\bZ}{\mathbf{Z}}

\newcommand{\balpha}{\boldsymbol{\alpha}}\newcommand{\bAlpha}{\boldsymbol{\Alpha}}
\newcommand{\bbeta}{\boldsymbol{\beta}}\newcommand{\bBeta}{\boldsymbol{\Beta}}
\newcommand{\bgamma}{\boldsymbol{\gamma}}\newcommand{\bGamma}{\boldsymbol{\Gamma}}
\newcommand{\bdelta}{\boldsymbol{\delta}}\newcommand{\bDelta}{\boldsymbol{\Delta}}
\newcommand{\bepsilon}{\boldsymbol{\epsilon}}\newcommand{\bEpsilon}{\boldsymbol{\Epsilon}}
\newcommand{\bzeta}{\boldsymbol{\zeta}}\newcommand{\bZeta}{\boldsymbol{\Zeta}}
\newcommand{\beeta}{\boldsymbol{\eta}}\newcommand{\bEta}{\boldsymbol{\Eta}} %
\newcommand{\btheta}{\boldsymbol{\theta}}\newcommand{\bTheta}{\boldsymbol{\Theta}}
\newcommand{\biota}{\boldsymbol{\iota}}\newcommand{\bIota}{\boldsymbol{\Iota}}
\newcommand{\bkappa}{\boldsymbol{\kappa}}\newcommand{\bKappa}{\boldsymbol{\Kappa}}
\newcommand{\blambda}{\boldsymbol{\lambda}}\newcommand{\bLambda}{\boldsymbol{\Lambda}}
\newcommand{\bmu}{\boldsymbol{\mu}}\newcommand{\bMu}{\boldsymbol{\Mu}}
\newcommand{\bnu}{\boldsymbol{\nu}}\newcommand{\bNu}{\boldsymbol{\Nu}}
\newcommand{\bxi}{\boldsymbol{\xi}}\newcommand{\bXi}{\boldsymbol{\Xi}}
\newcommand{\bomikron}{\boldsymbol{\omikron}}\newcommand{\bOmikron}{\boldsymbol{\Omikron}}
\newcommand{\bpi}{\boldsymbol{\pi}}\newcommand{\bPi}{\boldsymbol{\Pi}}
\newcommand{\brho}{\boldsymbol{\rho}}\newcommand{\bRho}{\boldsymbol{\Rho}}
\newcommand{\bsigma}{\boldsymbol{\sigma}}\newcommand{\bSigma}{\boldsymbol{\Sigma}}
\newcommand{\btau}{\boldsymbol{\tau}}\newcommand{\bTau}{\boldsymbol{\Tau}}
\newcommand{\bypsilon}{\boldsymbol{\ypsilon}}\newcommand{\bYpsilon}{\boldsymbol{\Ypsilon}}
\newcommand{\bphi}{\boldsymbol{\phi}}\newcommand{\bPhi}{\boldsymbol{\Phi}}
\newcommand{\bchi}{\boldsymbol{\chi}}\newcommand{\bChi}{\boldsymbol{\Chi}}
\newcommand{\bpsi}{\boldsymbol{\psi}}\newcommand{\bPsi}{\boldsymbol{\Psi}}
\newcommand{\bomega}{\boldsymbol{\omega}}\newcommand{\bOmega}{\boldsymbol{\Omega}}

\newcommand{\nA}{\mathbb{A}}
\newcommand{\nB}{\mathbb{B}}
\newcommand{\nC}{\mathbb{C}}
\newcommand{\nD}{\mathbb{D}}
\newcommand{\nE}{\mathbb{E}}
\newcommand{\nF}{\mathbb{F}}
\newcommand{\nG}{\mathbb{G}}
\newcommand{\nH}{\mathbb{H}}
\newcommand{\nI}{\mathbb{I}}
\newcommand{\nJ}{\mathbb{J}}
\newcommand{\nK}{\mathbb{K}}
\newcommand{\nL}{\mathbb{L}}
\newcommand{\nM}{\mathbb{M}}
\newcommand{\nN}{\mathbb{N}}
\newcommand{\nO}{\mathbb{O}}
\newcommand{\nP}{\mathbb{P}}
\newcommand{\nQ}{\mathbb{Q}}
\newcommand{\nR}{\mathbb{R}}
\newcommand{\nS}{\mathbb{S}}
\newcommand{\nT}{\mathbb{T}}
\newcommand{\nU}{\mathbb{U}}
\newcommand{\nV}{\mathbb{V}}
\newcommand{\nW}{\mathbb{W}}
\newcommand{\nX}{\mathbb{X}}
\newcommand{\nY}{\mathbb{Y}}
\newcommand{\nZ}{\mathbb{Z}}

\newcommand{\cA}{\mathcal{A}}
\newcommand{\cB}{\mathcal{B}}
\newcommand{\cC}{\mathcal{C}}
\newcommand{\cD}{\mathcal{D}}
\newcommand{\cE}{\mathcal{E}}
\newcommand{\cF}{\mathcal{F}}
\newcommand{\cG}{\mathcal{G}}
\newcommand{\cH}{\mathcal{H}}
\newcommand{\cI}{\mathcal{I}}
\newcommand{\cJ}{\mathcal{J}}
\newcommand{\cK}{\mathcal{K}}
\newcommand{\cL}{\mathcal{L}}
\newcommand{\cM}{\mathcal{M}}
\newcommand{\cN}{\mathcal{N}}
\newcommand{\cO}{\mathcal{O}}
\newcommand{\cP}{\mathcal{P}}
\newcommand{\cQ}{\mathcal{Q}}
\newcommand{\cR}{\mathcal{R}}
\newcommand{\cS}{\mathcal{S}}
\newcommand{\cT}{\mathcal{T}}
\newcommand{\cU}{\mathcal{U}}
\newcommand{\cV}{\mathcal{V}}
\newcommand{\cW}{\mathcal{W}}
\newcommand{\cX}{\mathcal{X}}
\newcommand{\cY}{\mathcal{Y}}
\newcommand{\cZ}{\mathcal{Z}}

\newcommand{\figref}[1]{Fig.~\ref{#1}}
\newcommand{\secref}[1]{Section~\ref{#1}}
\newcommand{\algref}[1]{Algorithm~\ref{#1}}
\newcommand{\eqnref}[1]{Eq.~\eqref{#1}}
\newcommand{\tabref}[1]{Table~\ref{#1}}

\def\mc{\mathcal}
\def\mb{\mathbf}

\newcommand{\T}{^{\raisemath{-1pt}{\mathsf{T}}}}

\newcommand{\Perp}{\perp\!\!\! \perp}

\makeatletter
\DeclareRobustCommand\onedot{\futurelet\@let@token\@onedot}
\def\@onedot{\ifx\@let@token.\else.\null\fi\xspace}
\def\eg{e.g\onedot} \def\Eg{E.g\onedot}
\def\ie{i.e\onedot} \def\Ie{I.e\onedot}
\def\cf{cf\onedot} \def\Cf{Cf\onedot}
\def\etc{etc\onedot}
\def\vs{vs\onedot}
\def\wrt{wrt\onedot}
\def\dof{d.o.f\onedot}
\def\etal{et~al\onedot}
\def\iid{i.i.d\onedot}
\def\evs{\emph{vs}\onedot}
\makeatother

\newcommand*\rot{\rotatebox{90}}

\newcommand{\boldparagraph}[1]{\vspace{0.4em}\noindent{\bf #1:}}

\definecolor{darkgreen}{rgb}{0,0.7,0}
\definecolor{lightred}{rgb}{1.,0.5,0.5}

\newcommand{\red}[1]{\noindent{\color{red}{#1}}}
\newcommand{\lightred}[1]{\noindent{\color{lightred}{#1}}}
\newcommand{\ag}[1]{ \noindent {\color{blue} {\bf Andreas:} {#1}} }
\newcommand{\lars}[1]{ \noindent {\color{cyan} {\bf Lars:} {#1}} }
\newcommand{\michael}[1]{ \noindent {\color{blue} {\bf Michael:} {#1}} }
\newcommand{\songyou}[1]{ \noindent {\color{red} {\bf Songyou:} {#1}} }

\begin{abstract}

Recently, neural implicit surfaces have become popular for multi-view reconstruction. To facilitate practical applications like scene editing and manipulation, some works extend the framework with semantic masks input for the object-compositional reconstruction rather than the holistic perspective. Though achieving plausible disentanglement, the performance drops significantly when processing the indoor scenes where objects are usually partially observed. We propose RICO to address this by regularizing the unobservable regions for indoor compositional reconstruction. Our key idea is to first regularize the smoothness of the occluded background, which then in turn guides the foreground object reconstruction in unobservable regions based on the object-background relationship. Particularly, we regularize the geometry smoothness of occluded background patches. With the improved background surface, the signed distance function and the reversedly rendered depth of objects can be optimized to bound them within the background range. Extensive experiments show our method outperforms other methods on synthetic and real-world indoor scenes and prove the effectiveness of proposed regularizations.
The code is available at \href{https://github.com/kyleleey/RICO}{https://github.com/kyleleey/RICO}

\end{abstract}


\section{Introduction}
\label{intro}

\begin{figure}
\begin{center}
\includegraphics[width=\linewidth]{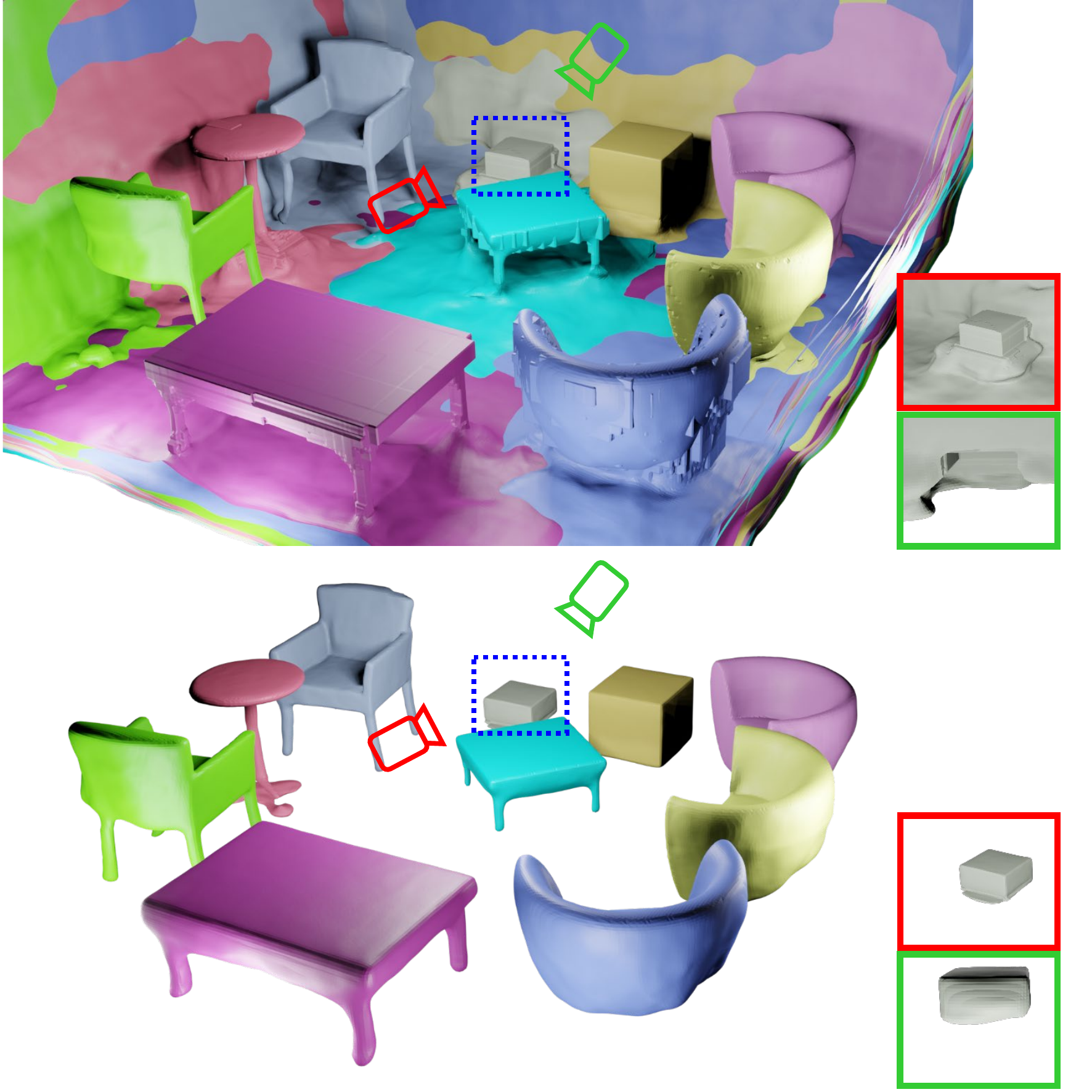}
\end{center}
\caption{\textbf{Comparison} of~\cite{wu2022object}~(Top) and ours~(Bottom). Different objects are visualized in different colors. Previous reconstructions are connected with large artifacts. Taking the partially observed cubic object's front and back views~(in red and green rectangles respectively) as an example, \cite{wu2022object} can only get the visible surface, while ours can complement the complete body.}
\label{fig:intro}
\end{figure}

Reconstructing 3D geometry from images is a fundamental problem in computer vision and has many downstream applications like VR/AR and game assets creation. With the advance of neural implicit representations~\cite{mildenhall2021nerf}, recent reconstruction methods~\cite{oechsle2021unisurf,wang2021neus,yariv2021volume,yu2022monosdf} can recover accurate geometry from multi-view images.
However, existing methods typically regard the whole scene as an entirety and reconstruct everything altogether, thus preventing applications like scene editing. In indoor scenes with plenty of reconfigurable objects, a disentangled object-compositional reconstruction, \ie decomposing the geometry into different instantiated objects and background, can be more suitable for further applications like moving the sofa in the scene.

\begin{figure}
\begin{center}
\includegraphics[width=\linewidth]{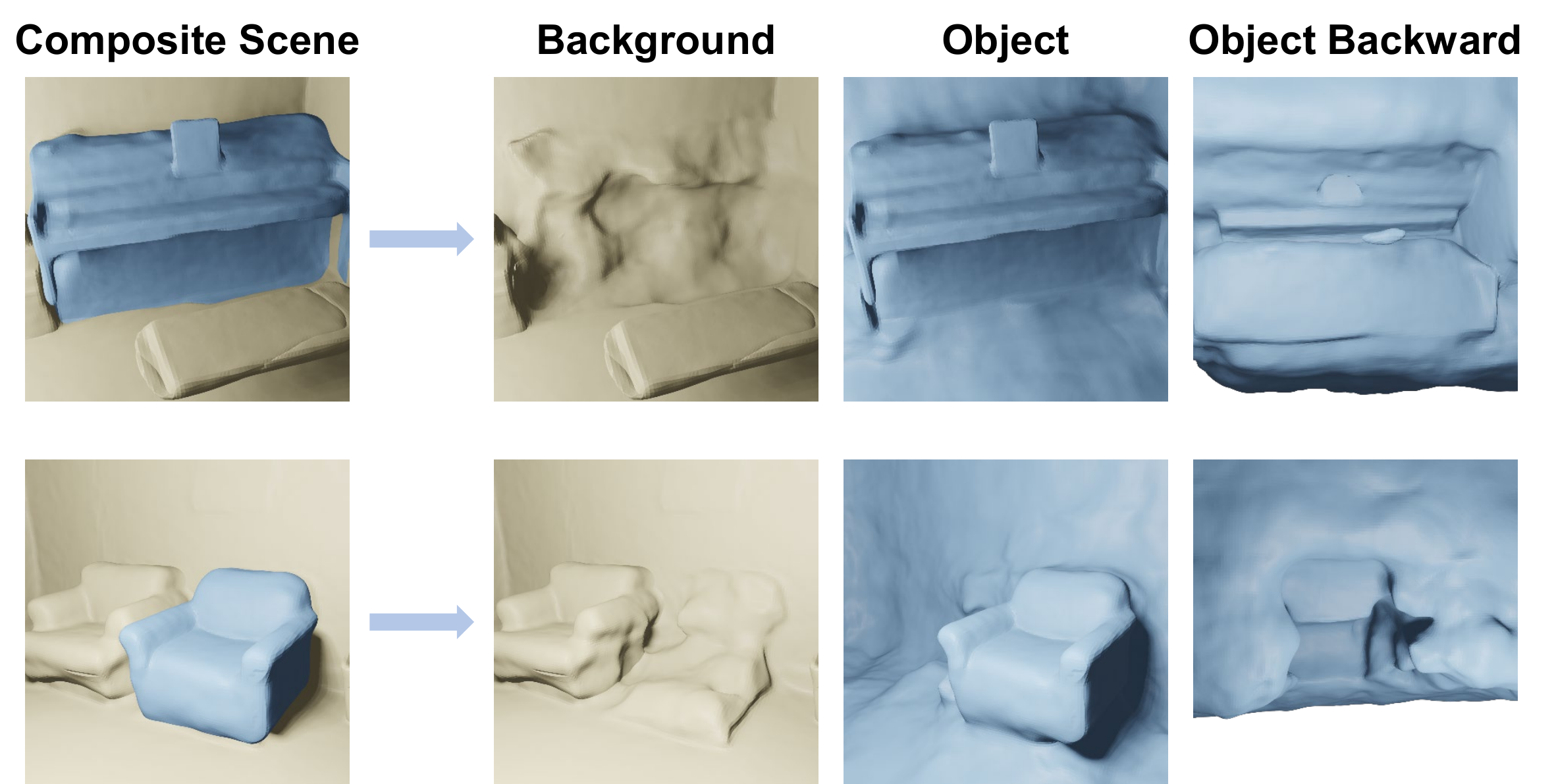}
\end{center}
   \caption{\textbf{ObjSDF results.} Interested objects are dyed in blue. Despite of the plausible composition, the disentangled backgrounds have artifacts and sunk holes, and partially observed objects can only get the visible surface~(illustrated in `Object Backward').}
\label{fig:intro-objsdf-edit}
\end{figure}

In this paper, we aim to recover the room-level indoor scenes with decomposed geometry of individual objects and background.
We assume that multi-view posed images and semantic masks that assign different labels to each instantiated object and the background are given as input.
Existing object-compositional methods~\cite{guo2020object,zhi2021place,yang2021learning} concentrate more on the rendering performance rather than the underlying geometry, and thus can not be directly used for reconstruction. 
The most recent work ObjSDF~\cite{wu2022object} learns an object-compositional signed distance function~(SDF) field by proposing a transform function between SDF values and semantic logits. Specifically speaking, ObjSDF predicts multiple SDF values at each 3D point for different semantic classes, and converts them to semantic logits, allowing for separating object SDF values from the background when supervised by semantic masks. 
Although achieving plausible shape disentanglement, it suffers from a common problem in indoor scenes: objects and background can only be \textit{partially observed due to occlusions}. When the object is partially observed, \eg a cubic object against the wall, ObjSDF can not properly reconstruct the geometry between them~(see Fig.~\ref{fig:intro} top-right). The reason is that the existing works~\cite{zhi2021place,wu2022object} can only effectively regularize semantic labels and geometry of observed regions, and have little impact on the unobserved regions. 
When processing the indoor scenes where a large portion of objects are partially observed, the reconstruction results of these objects will be visible surface connected with the unconstrained structures~(as shown in Fig.~\ref{fig:intro}, see Section~\ref{method-obj-scene} for a detailed analysis).
Fig.~\ref{fig:intro-objsdf-edit} shows that even with reasonable reconstruction when composing all the objects together, each object's result in the unobserved region is far from satisfactory and can hinder further applications like manipulating the object.

We propose \netname, which realizes the proper geometry disentanglement for indoor scenes~(see Fig.~\ref{fig:intro} bottom) by explicitly regularizing the unobserved regions. To be more specific, when the object is partially observed, recovering its geometry is an ill-posed problem even with corresponding masks. Thus, introducing prior regularization for unobserved regions is necessary. We exploit two types of prior knowledge for indoor scenes in this work: 1)~background smoothness and 2)~object-background relations. First, when one ray hits the object surface, the existing method~\cite{wu2022object} can properly regularize the geometry and appearance on the hitting point, but can not account for the background surface behind this object. This drawback leads to artifacts and holes on the unobserved background surface~(see Fig.~\ref{fig:intro-objsdf-edit}). We propose a patch-based smoothness loss to regularize the SDF values of unobserved background regions. Then, since the background reconstruction is improved, we can leverage another strong prior: \textit{the objects are all within the room}, \ie using the background surface to regularize the SDF field of objects. We design two regularization terms: an object point-SDF loss for sampled points behind the background surface and a reversed depth loss to regularize the SDF distribution of the entire ray. Both terms aim to bound the object within the background surface's range, thus preventing the aforementioned unmeaning structure, making the object reconstruction a \textit{watertight and plausible} shape instead of an open surface with severe artifacts.

In summary, we propose RICO to realize compositional reconstruction in indoor scenes where a large portion of objects are partially observed. Our main contributions are: 
i)~A patch-based background smoothness regularizer for unobserved background surface geometry. ii)~Guided by the improved background surface, we exploit the object-background relation as prior knowledge and design objectives that effectively regularize objects' unobserved regions. iii)~Extensive experiments on both real-world and synthetic datasets prove our superior reconstruction performance compared to previous works, especially for the partially observed objects.

\section{Related Work}
\label{related}

\begin{figure*}[htbp]
\begin{center}
\includegraphics[width=0.95\linewidth]{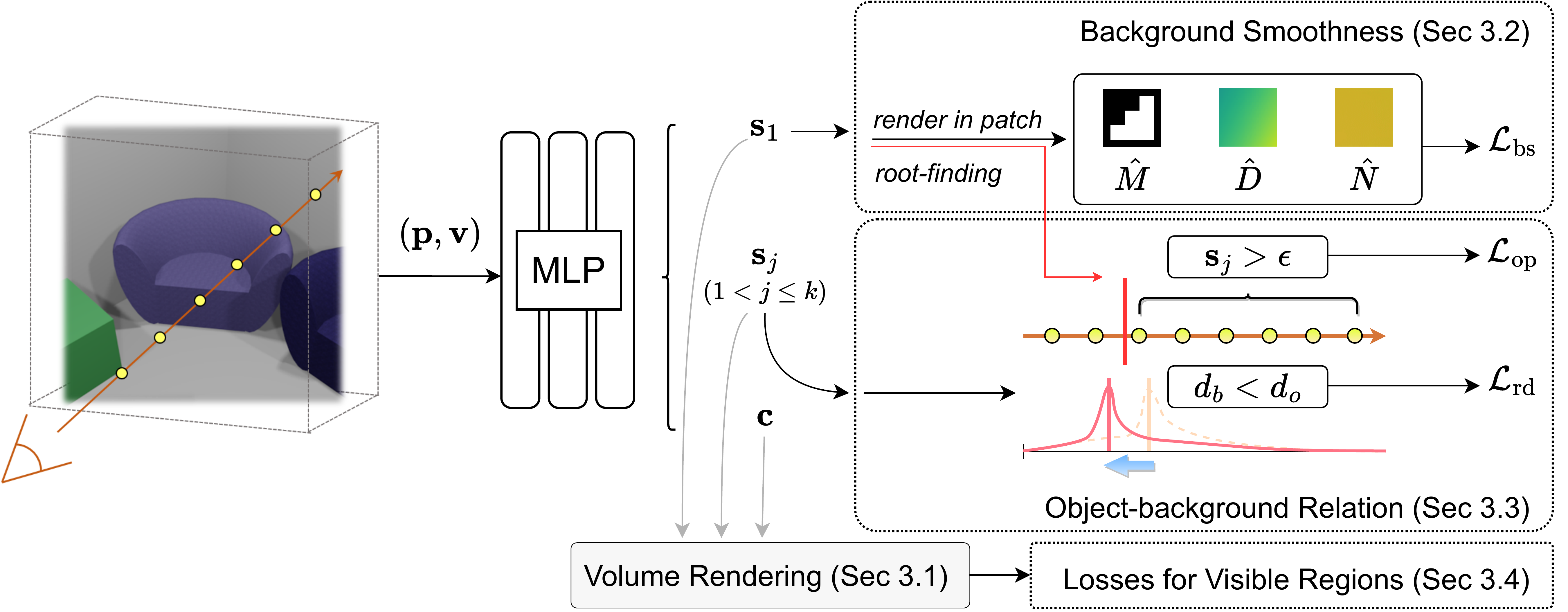}
\end{center}
   \caption{\textbf{Overview.} In this work, we propose two different regularizations. We first regularize the geometry smoothness of the unobserved background regions in a sampled patch. Then, we exploit the background surface as the prior to constrain the objects' surface. In detail, a per-point SDF loss and a reversed depth loss are introduced to regularize the manifold of objects' SDF functions. Combined with other reconstruction losses, our method reaches a neat and disentangled compositional reconstruction in indoor scenes.}
\label{fig:method}
\end{figure*}

\boldparagraph{Neural Implicit Representation for Reconstruction}
\label{related-inr-recon}
Recently,
neural implicit representations have emerged as a popular representation for 3D reconstruction.
While early approaches~\cite{mescheder2019occupancy,park2019deepsdf} rely on 3D supervision, a few works~\cite{niemeyer2020differentiable,yariv2020multiview} exploit surface rendering to use multi-view image supervision only, but also suffer from the unstable training. Neural radiance field~(NeRF)~\cite{mildenhall2021nerf} adopts volume rendering technique to achieve photorealistic scene representation and stable optimization. However, NeRF's formulation can not guarantee accurate underlying geometry. 
Therefore, \cite{oechsle2021unisurf,wang2021neus,yariv2021volume} combine the geometry representation with iso-surface ~(\eg occupancy~\cite{mescheder2019occupancy}, SDF~\cite{wang2021neus,yariv2021volume}) and volume density to accurately reconstruct object-level scenes from RGB images.
\cite{guo2022neural} further applies the planar regularization for scene-level reconstruction. To tackle the problem in texture-less regions, \cite{wang2022neuris,yu2022monosdf} utilize results from pretrained normal and depth estimation networks to guide the SDF training and boost the reconstruction performance.

However, despite the promising reconstruction performance, the aforementioned methods all consider the whole scene as an entirety. Our method focuses on decomposing the scene reconstruction into the background and different foreground objects, which can be regarded as compositional scene reconstruction.

\boldparagraph{Compositional Scene Reconstruction}
\label{related-composite-recon}
Decomposing a scene into its different components could benefit downstream applications like scene editing. Many works have been proposed to recover the scene in a compositional manner from different perspectives. \cite{nie2020total3dunderstanding,zhang2021holistic,irshad2022shapo,nie2022learning} detect and reconstruct different objects in the given monocular image, and predict the scene's layout at the same time. But most of these methods require large-scale datasets with 3D ground truth for training. \cite{kobayashi2022decomposing,tschernezki2022neural,mazur2022feature} optimize a feature field from a large pretrained model~\cite{caron2021emerging,li2022language}, which enables deep feature based decomposition and manipulation.
\cite{wu2022d,sharma2022seeing} exploit the self-supervise paradigm to decompose the scenes into static and dynamic parts. 
More works~\cite{guo2020object,zhi2021place,yang2021learning,yang2022neural,wang2022dm,fu2022panoptic,kundu2022panoptic,wallingford2023neural} focus on recovering the object-compositional scenes given semantic masks with images. However, this line of works concerns the rendering outputs rather than the geometry, \ie the reconstruction results are sub-optimal.

Recently, ObjSDF~\cite{wu2022object} proposes a transform function between semantic logits and SDFs of different objects, which enables optimizing SDF fields with image and semantic mask supervision, and decomposing the whole scene with accurate reconstruction. However, in the indoor scenes where many objects are in partial observation, its reconstruction results are far from satisfactory and can not be used for further applications~(see Fig.~\ref{fig:intro-objsdf-edit}). On the contrary, our method introduces geometry prior to unobserved regions, yielding better compositional scene reconstruction.

\boldparagraph{Prior Regularization in Neural Implicit Reconstruction}
\label{related-prior}
In addition to the commonly used RGB image supervision, many different priors have been proposed to benefit the neural implicit representation. 
For example, \cite{deng2022depth,roessle2022dense} use explicit point cloud as the depth prior, \cite{niemeyer2022regnerf} employs regularization in unseen views and \cite{jain2021putting} introduces pretrained model's feature consistency.
\cite{jiang2022neuman,pavlakos2022one} adopt explicit human model as the structural prior for the human-centric scenes. 
As for surface reconstruction, \cite{guo2022neural} proposes a manhattan-world assumption for the planar regions, \cite{yu2022monosdf} utilizes the normal and depth prediction from off-the-shelf model~\cite{eftekhar2021omnidata} as prior 
to regularize the texture-less regions.

Our method exploits the geometry prior in the unobserved region for compositional scene reconstruction. The proposed regularization can effectively reconstruct the objects and disentangle them from the background, even if the object itself is partially observed.

\section{Methodology}
\label{method}
Our goal is to recover decomposed geometry surface of the objects and background within a scene from the images and semantic masks inputs. To this end, we first review the SDF-based neural implicit representation and how to use the semantic logits for compositional reconstruction in Section~\ref{method-background}. Next, we propose two types of regularizations on unobserved regions to address the partial observation problem: patch-based background smoothness~(Section~\ref{method-bgloss}) and object-background relation~(Section~\ref{method-obj-scene}). Finally, we introduce the overall optimization procedure in Section~\ref{method-train}. An overview of our method is provided in Fig~\ref{fig:method}.

\subsection{Background}
\label{method-background}

\boldparagraph{Volume Rendering of SDF-based Implicit Surface}
For implicit reconstruction, the geometry of the scene is represented as the signed distance function~(SDF) $s(\bp)$ of each spatial point $\bp$, which is the point's distance to the closest surface. In practice, the SDF function is implemented as a multi-layer perceptron~(MLP) network $f(\cdot)$. The appearance of the scene is also defined as an MLP $g(\cdot)$:
\begin{equation}
    \begin{aligned}
        f & : \bp\in\nR^3 \mapsto (s\in\nR,\bff\in\nR^{256}) \\
        g & : (\bp\in\nR^3, \bn\in\nR^3, \bv\in\nS^2, \bff\in\nR^{256}) \mapsto \bc\in\nR^3
    \end{aligned}
\end{equation}
where $\bff$ is a geometry feature vector, $\bn$ is the normal at $\bp$, $\bv$ is the viewing direction and $\bc$ is the view-dependent color.
We adopt the unbiased rendering proposed in \cite{wang2021neus} to render the image. 
For each camera ray $\br=(\bo, \bv)$ with $\bo$ as the ray origin, $n$ points $\{\bp(t_i) = \bo + t_i\bv|i=0,1,\dots,n-1\}$ are sampled, and the pixel color can be approximated as: 
\begin{equation}
    \hat{\bC}(\br) = \sum_{i=0}^{n-1}T_i\alpha_i\bc_i.
    \label{eq-volume-rendering}
\end{equation}
The $T_i$ is the discrete accumulated transmittance derived from $T_i=\prod_{j=0}^{i-1}(1-\alpha_j)$, and $\alpha_i$ is the discrete density value defined as 
\begin{equation}
    \alpha_i = \max\left(\frac{\Phi_u(s(\bp(t_i))) - \Phi_u(s(\bp(t_{i+1})))}{\Phi_u(s(\bp(t_i)))}, 0\right),
\end{equation}
where $\Phi_u(x) = (1+e^{-ux})^{-1}$  and $u$ is a learnable parameter. By minimizing the difference between predicted and ground-truth pixel colors, we can learn the SDF and appearance function of the desired scene.

\boldparagraph{Learning SDF with Semantic Logits}
In this work, we consider compositional reconstruction of $k$ objects given their masks. Note that we also consider the background as an instantiated object for brevity as in \cite{wu2022object} and follow their network structure.
In detail, for a scene with $k$ objects, the SDF MLP $f(\cdot)$ now outputs $k$ SDFs at each point, and the $j$-th~($1\leq j\leq k$) SDF represents the geometry of $j$-th object. Without loss of generality, we set $j=1$ as the background category and others for objects in Fig.~\ref{fig:method} and the rest of the paper.
The \textit{scene} SDF is the minimum of $\{s_j\}$, which is used for sampling points along the ray and aforementioned volume rendering~(Eq.~\ref{eq-volume-rendering}). Moreover, each point's $k$ SDFs can be transformed into semantic logits $\bh(\bp)$ as
\begin{equation}
    \begin{aligned}
        h_j(\bp) &= \gamma / (1 + \exp(\gamma\cdot s_j(\bp))), \\
        \bh(\bp) &= [h_1(\bp), h_2(\bp),...,h_k(\bp)],
    \end{aligned}
\end{equation}
where $\gamma$ is a fixed parameter. Using volume rendering to accumulate the semantic logits of all the points along a ray, we can get the semantic logits $\bH(\br)\in \nR^k$ of each pixel. During training, the cross-entropy loss applied to $\bH(\br)$ is backpropagated to the SDF values, allowing for learning the compositional geometry.

\subsection{Patch-based Background Smoothness}
\label{method-bgloss}

Although the volume rendering can propagate gradients along the entire ray, the optimization mainly focuses on the surface-hitting point, as its accumulated weight can be much larger than the others. As a result, the geometry of the points behind the first-hit surface can not be optimized correctly. In the indoor scenes, the occluded part of the background surface is invisible in all the images, 
which
can be with holes and random artifacts~(see Fig.~\ref{fig:intro-objsdf-edit}).

Since we cannot tell the exact color, depth or normal of the occluded part, it is intractable to optimize this region \wrt its ground truth. Therefore, we propose to regularize the geometry of the occluded background to be \textit{smooth}, thus preventing some clearly wrong artifacts.

In detail, we regularize the smoothness of rendered depth and normal of background surface within a small patch region. To save the computation budget, we randomly sample a $\cP\times \cP$ size patch every $\cT_{\cP}$ iterations in the given image and sample points along the patch rays using the \textit{background SDF} only. 
We compute depth map $\hat{D}(\br)$ and normal map $\hat{N}(\br)$ of the sampled patch following \cite{yu2022monosdf}, $\br$ denotes the sampled ray in patch. The semantic map of the patch is also computed and transformed into a patch Mask $\hat{M}(\br)$:
\begin{equation}
    \hat{M}(\br) = \mathbbm{1}[\arg\max (\bH(\br)) \neq 1],
\end{equation}
which means the mask value is $1$ if the rendered class is not the background, so that only the occluded background is regularized. Taking rendered depth as an example, the patch-based background smoothness loss is
\begin{equation}
\label{eq-smooth-depth-loss}
    \begin{aligned}
        \cL(\hat{D}) =& \sum_{d=0} ^3 \sum_{m,n=0}^{\cP-1-2^d} \hat{M}(\br_{m,n}) \odot (|\hat{D}(\br_{m,n}) - \\
        &\hat{D}(\br_{m,n+2^d})| + |\hat{D}(\br_{m,n}) - \hat{D}(\br_{m+2^d,n})| ).
    \end{aligned}
\end{equation}
Here the smoothness is applied on different intervals controlled by $d$. $m$ and $n$ are the pixel indices within the patch and the mask is multiplied at each position with hadamard product $\odot$. The normal smoothness loss $\cL(\hat{N})$ can be obtained similarly. We define the overall background smoothness loss $\cL_{bs}$ as:
\begin{equation}
    \cL_{\text{bs}} = \cL(\hat{D}) + \cL(\hat{N}).
\end{equation}
Here in contrast to \cite{niemeyer2022regnerf} which applies a patch-based regularization to visible regions in other views, we instead regularize the \textit{occluded} regions of the background.

\begin{figure}[!htbp]
\begin{center}
\includegraphics[width=\linewidth]{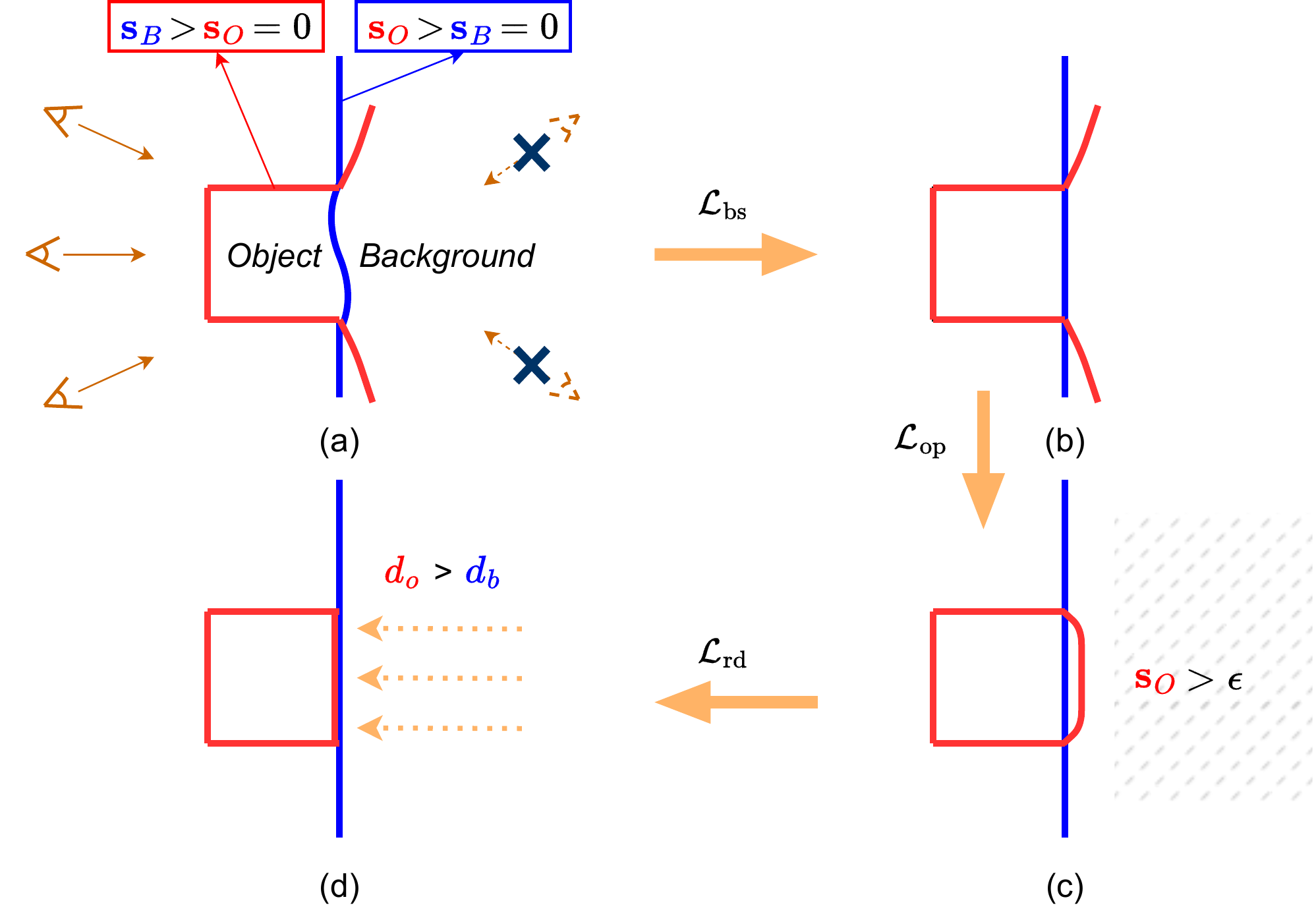}
\end{center}
   \caption{\textbf{Toy-case analysis.} A bird-eye view of an object against the background is shown here, the SDF and surface of object is shown in red and background's in blue. (a)~In \cite{wu2022object} the minimum SDF is used for volume rendering and semantic loss, so on object's visible surface where object SDF is optimized to $0$ and the background SDF here is positive, and similar for the visible background surface. Since the scene is partially observed from left, the right part of object surface is open and unobserved background region is unconstraint. (b)~With the smoothness prior the background surface can be plausible, (c)~and object point SDF loss close the object surface but still have intersections, (d)~finally the reversed depth loss optimizes the entire ray thus the object can be within the background.}
\label{fig:obj-case-vis}
\end{figure}

\subsection{Object-background Relation as Regularization}
\label{method-obj-scene}

With the help of the patch-based background smoothness loss, most artifacts of the background are resolved, yielding a smooth surface. We further leverage this smooth background surface to regularize the SDF fields of other objects, leveraging a key prior knowledge that \textit{all the other objects are confined to the room}, \ie, the background surface.

In the original framework of \cite{wu2022object}, if an object is partially observed, \eg against the background, the object's reconstructed surface won't be a ``closed" surface~(see Fig.~\ref{fig:intro} and Fig.~\ref{fig:intro-objsdf-edit}).
We refer to the toy-case analysis shown in Fig.~\ref{fig:obj-case-vis} for the reason of the current unsatisfactory reconstruction.
To encourage the reconstruction to be \textit{watertight}, we design two types of regularization on the SDF fields of objects.

\boldparagraph{Object Point-SDF Loss}
A straightforward solution is to directly regularize the objects' SDFs on every sampled point behind the background surface. To be more specific, the regular volume rendering only guarantees a single change of the sign (positive SDF to negative) when a ray hits a visible surface. For watertight objects, the ray should hit another occluded surface (negative SDF to positive). With the prior that the object is confined within the room, the occluded object surface should be closer than the background surface, meaning that the object SDFs of points behind the background surface should all be positive~(Fig.~\ref{fig:obj-case-vis}~(c)).

We implement an object point-SDF loss based on the above analysis. For the sampled points along the rays, we first utilize the root-finding algorithm among the background SDF of these points and find the zero-SDF ray depth $t'$. Then the object point-SDF loss can be formulated as
\begin{equation}
    \cL_{\text{op}} = \frac{1}{k-1}\sum_{j=2}^{k}\max\left(0, \epsilon - \bs_j(\bp(t_i))\right)\cdot\mathbbm{1}[t_i > t'] ,
\end{equation}
which pushes the objects' SDFs at points behind the surface larger than a positive threshold $\epsilon$.

\boldparagraph{Reversed Depth Loss}
Although the $\cL_{\text{op}}$ can effectively regularize the SDF fields of each object, in practice we find the reconstructed object surface can still have intersections with the background surface~(Fig.~\ref{fig:obj-case-vis}~(c)). The reason is that the sampled points are discrete and in most cases, the background surface is between two sampled points. Therefore, the sign change of the occluded surface may still occur after hitting the background surface.

Since per-point optimization can not effectively propagate to the distribution of the entire ray. To optimize the entire ray's SDF distribution for better regularization, we compute a \textit{reversed depth} along each ray. With the help of $\cL_{\text{op}}$, the sign of the object SDF along one ray now is positive-negative-positive, which enables rendering a depth value backward. We first transform the ray depth $\{t_i|i=0,1,\dots,n-1\}$ into the reversed ray depth named $\{\hat{t}_i|i=0,1,\dots,n-1\}$, where
\begin{equation}
    \hat{t}_i = (t_0 + t_{n-1}) - t_{n-1-i}.
\end{equation}
With the reversed ray depth values, we use the background and object SDF both in reversed order to compute the accumulated weight and get the reversed depth respectively. Remarkably, in order to compute the exact correct depth, the points should be re-sampled along the reverse direction. 
Here we directly use the sampled points to avoid computation overhead, and empirical results prove its effectiveness. We only compute the reverse depth of one pixel if satisfying two conditions: 1)~this pixel's $\hat{M}(\br)=1$; 2)~the SDF value of the rendered object at the furthest point is positive. Note that the second condition is usually satisfied when the object point-SDF loss is applied. By computing the reversed depth $d_o$ of the hitting object~(determined by the pixel's rendered semantic) and $d_b$ of the background, we can get the reversed depth loss:
\begin{equation}
    \cL_{\text{rd}} = \max(0, d_b - d_o),
\end{equation}
which pushes the object surface within the background as illustrated in Fig.~\ref{fig:obj-case-vis}~(d).

\subsection{Training Objectives Details}
\label{method-train}

Monocular geometric cues are essential for indoor scene reconstruction as proved in~\cite{yu2022monosdf}. Following \cite{yu2022monosdf}, we add the depth and normal consistency loss~($\cL_{\text{D}}$,$\cL_{\text{N}}$) with pseudo ground truth from Omnidata~\cite{eftekhar2021omnidata} model. In experiments, we also add the monocular cues on \cite{wu2022object} to get a stronger baseline for a fair comparison.

The SDF network is also regularized by an Eikonal~\cite{gropp2020implicit} loss item $\cL_{\text{E}}$. We further use the semantic loss $\cL_{\text{S}}$ proposed in \cite{wu2022object} to learn the compositional geometry. The overall loss function for compositional reconstruction is:
\begin{equation}
    \begin{aligned}
        \cL =& \cL_{\text{RGB}} + \lambda_{\text{D}}\cL_{\text{D}} + \lambda_{\text{N}}\cL_{\text{N}} + \lambda_{\text{E}}\cL_{\text{E}} + \lambda_{\text{S}}\cL_{\text{S}} \\
    &+ \lambda_{\text{bs}}\cL_{\text{bs}} + \lambda_{\text{op}}\cL_{\text{op}} + \lambda_{\text{rd}}\cL_{\text{rd}}.
    \end{aligned}
\end{equation}
We set $\lambda_{\text{bs}},\lambda_{\text{op}},\lambda_{\text{rd}}=0.1$ in our experiments. 
For other loss terms, given the weight of RGB reconstruction loss as $1$, we set $\lambda_{\text{D}} = 0.1$ and $\lambda_{\text{N}} = \lambda_{\text{E}} = 0.05$ following \cite{yu2022monosdf}. For the semantic loss, we follow the implementation in \cite{wu2022object} and set $\lambda_{\text{D}} = 0.04$.
The detailed calculation of other losses can be found in supplementary.



\section{Experiments}
\label{experiments}
 
We first conduct ablation study on each proposed component. 
Then we provide quantitative and qualitative results of object-compositional reconstruction on real-world and synthetic scenes.
Finally, we show some possible object manipulation with our compositional reconstruction.

\boldparagraph{Datasets}
We consider three types of indoor datasets with multi-view RGB images and masks: 
1)~ScanNet~\cite{dai2017scannet}, a real-world dataset widely used in previous works~\cite{guo2022neural,wang2022neuris,wu2022object,yu2022monosdf}; 
2)~ToyDesk dataset, a real-world dataset with two scenes and each has four objects placed on the table plane. This dataset has also been widely used in previous compositional works~\cite{yang2021learning,wu2022object}
3)~a hand-crafted synthetic dataset with five scenes, each containing $5\sim 10$ objects. The synthetic dataset is considered such that the ground truth geometry of both, occluded and non-occluded regions are available.

\boldparagraph{Metrics}
For reconstruction performance, we report Chamfer Distance, F-score for evaluation on ScanNet, following \cite{guo2022neural,yu2022monosdf}. On synthetic scenes we divide metrics for two aspects: objects and background. For objects we report the reconstruction performance compared to the complete ground truth object mesh, and the final results are averaged across all the objects in all the scenes. For background we report the reconstruction metrics and rendered depth errors of the occluded regions only to highlight the effectiveness of our regularization. The results are also averaged across all the scenes. See supplementary for a detailed introduction to datasets and metrics.
We also report two 2D metrics, PSNR and mIoU, on the ToyDesk and ScanNet scenes. Here we omit the synthetic dataset considering its relatively simple texture. For the test image set, on ToyDesk dataset we use the official split, and on ScanNet we sample frames from their original camera trajectories. The metrics are averaged across different images of different scenes.

\begin{figure}
\begin{center}
\includegraphics[width=0.9\linewidth]{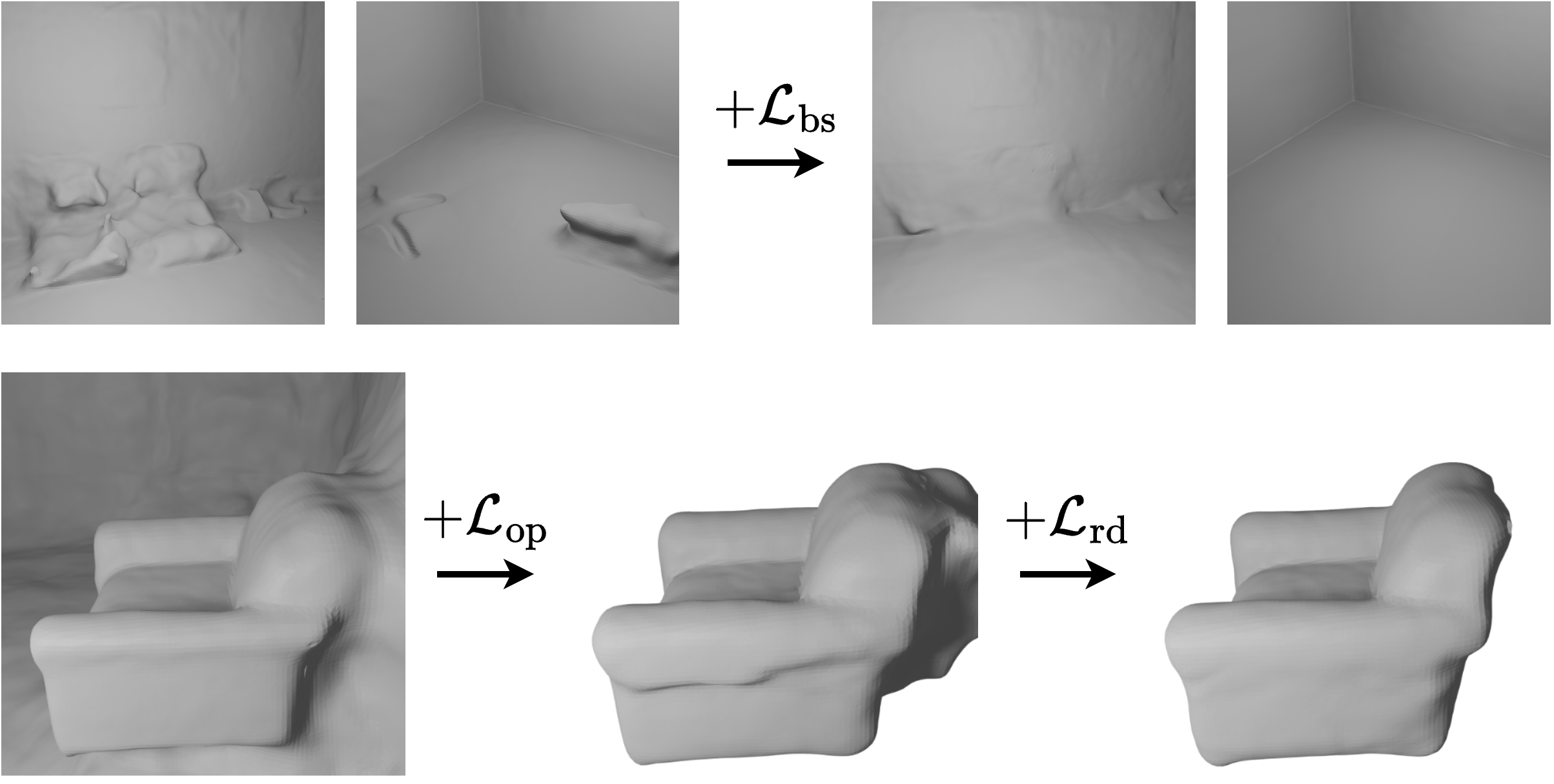}
\end{center}
   \caption{\textbf{Effects of Regularizations.} The first row shows two backgrounds, the $\cL_{\text{bs}}$ clearly mitigates most of the artifacts. The last row shows that with $\cL_{\text{op}}$ we can get the watertight object and $\cL_{\text{rd}}$ further constrains the unobserved surface.}
\label{fig:ablation}
\end{figure}

\begin{figure*}[t]
\begin{center}
\includegraphics[width=0.95\linewidth]{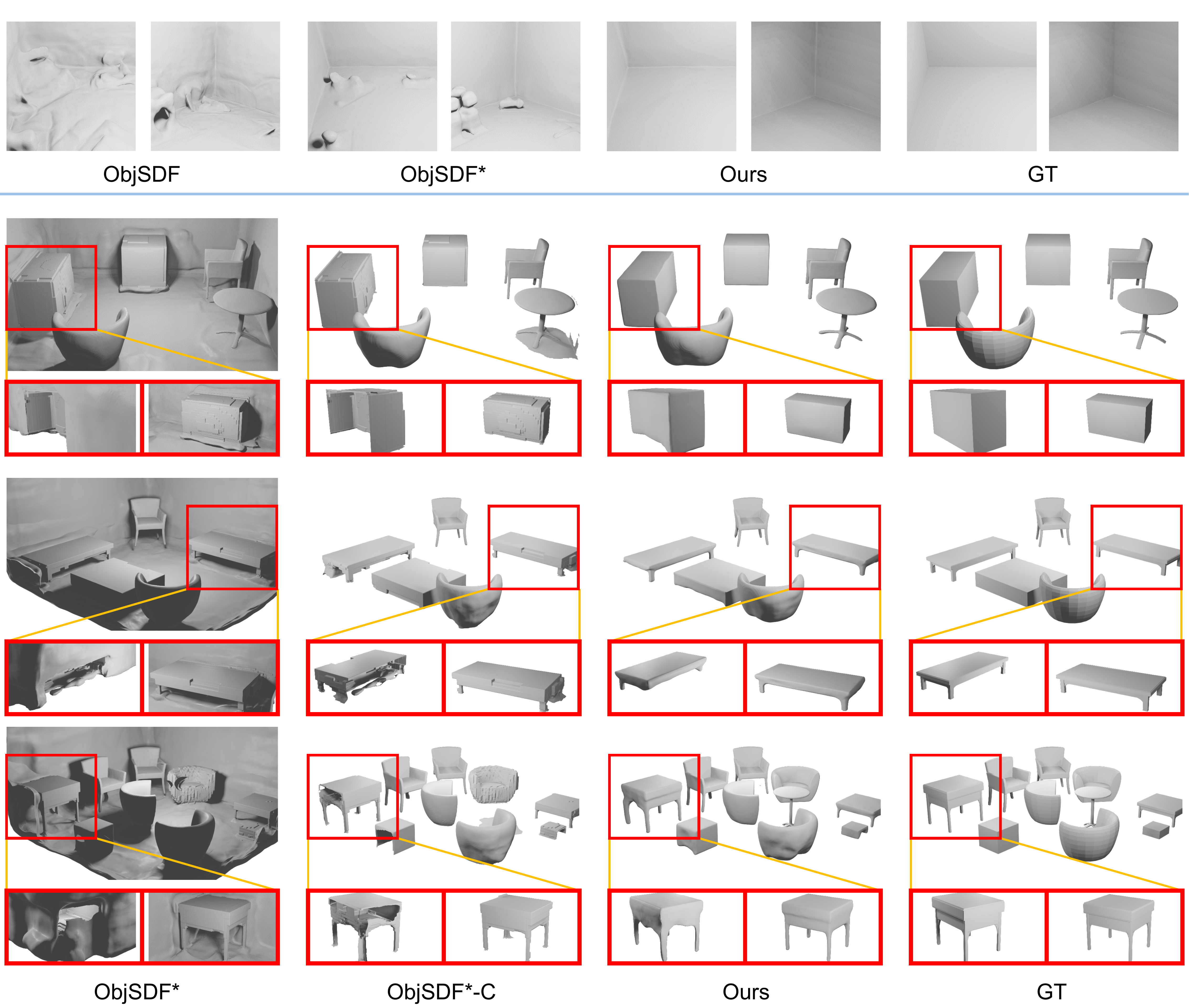}
\end{center}
   \caption{\textbf{Qualitative Results on Synthetic Scenes.} Above the blue line we show the comparison of different methods on two background scenes. Below we provide results of two scenes where only the object results are shown. In red rectangles at the bottom of each picture, we show the back~(left part) and front~(right part) views of an example object. Detailed descriptions in Section~\ref{exp:synthetic-reconstruction}.}
\label{fig:synthetic}
\end{figure*}

\begin{figure*}[t]
\begin{center}
\includegraphics[width=0.95\linewidth]{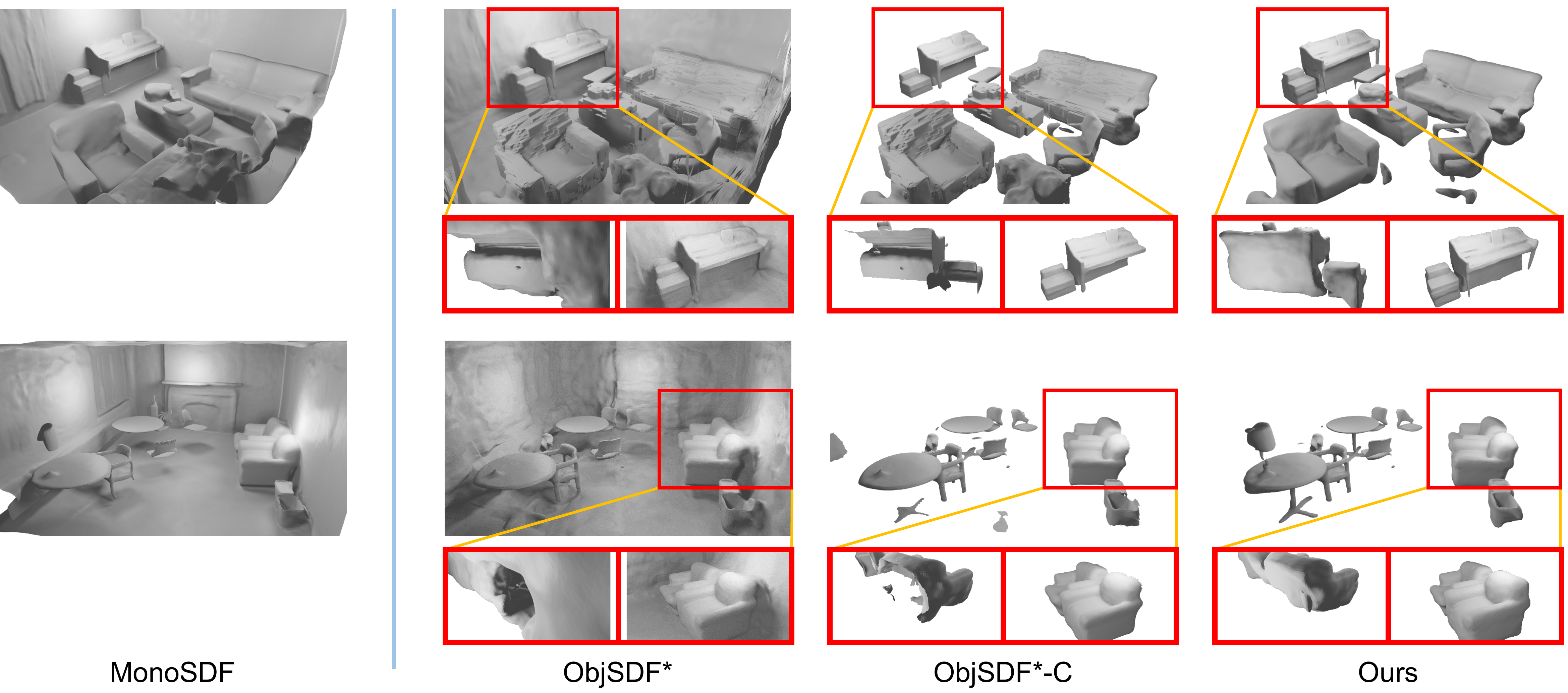}
\end{center}
   \caption{\textbf{Qualitative Results on ScanNet~\cite{dai2017scannet}.} On the left of blue line we show the overall reconstruction from \cite{yu2022monosdf} as the reference. On the right we show the comparison between our method and two baselines. Similarly, the back and front views of objects with partial observations are provided in red rectangles. Detailed descriptions in Section~\ref{exp-scannet}.}
\label{fig:scannet}
\end{figure*}

\boldparagraph{Baselines}
We mainly compare with \textbf{ObjSDF}~\cite{wu2022object} in this work, as it is the only method that focuses on the same task, a specific discussion is provided in the supplementary. Since we focus on the indoor scene where monocular cues can benefit a lot~\cite{yu2022monosdf}, we add losses proposed in \cite{yu2022monosdf} to our method for better performance. For a fair comparison, we also combine \cite{wu2022object} and \cite{yu2022monosdf} for a stronger baseline, named \textbf{ObjSDF*}. 
Moreover, since ObjSDF*'s object reconstructions inevitably have artifacts, we develop a post-process method to cull the parts outside the background range and name this baseline \textbf{ObjSDF*-C}. We provide the details of how to set the range in the supplementary. Note that our method directly generates clean watertight meshes and doesn't need such post-processing. In ScanNet experiments, we also report results for MonoSDF~\cite{yu2022monosdf} as a reference since we can only evaluate the overall reconstruction~(details in Section~\ref{exp-scannet}).

\boldparagraph{Implementation Details}
We implement our method in PyTorch~\cite{paszke2019pytorch}. We use the Adam~\cite{kingma2014adam} optimizer with a learning rate of 5e-4 for 50k iterations and sample 1024 rays per iteration. The weight initialization scheme for SDF MLP is identical to \cite{yariv2021volume,wang2021neus,wu2022object}. The $u$ is initialized as $0.05$ and we set $\gamma$=$20$ as proposed in \cite{wu2022object}. $\cP$, $\cT_{\cP}$ and $\epsilon$ are set as $32$, $10$ and $0.05$ respectively. All the reconstructions are acquired by using marching cube~\cite{lorensen1987marching} at the resolution of $512$. 

Moreover, we adopt the geometric initialization~\cite{gropp2020implicit} for the geometry network, which initializes the reconstruction with a unit sphere and the surface normals are facing inside at the beginning of the optimization. For the ScanNet dataset, we follow the protocol in \cite{yu2022monosdf} to crop the input image to $384\times 384$ and adjust the intrinsics accordingly. For the synthetic dataset, we directly render the image at the same resolution. Since we focus on the indoor scenes in this paper, we adopt the common practice to process the ``bounded'' scene. For each scene, we normalize the camera poses so that all the cameras lie in a unit sphere. The rays intersection, \ie the furthest sampling location, is computed based on this sphere and we also conduct Marching Cubes~\cite{lorensen1987marching} within the same area for the final reconstruction.

\subsection{Ablation Study}
\label{section-ablation}

\begin{table}[t]
	\centering
	\resizebox{0.99\linewidth}{!}{%
		\begin{tabular}{lccc|ccc|cc}
\toprule
     & & & & \multicolumn{3}{c}{Background} & \multicolumn{2}{c}{Object} \\
     \midrule
     & $\cL_{\text{bs}}$ & $\cL_{\text{op}}$  & $\cL_{\text{rd}}$ & D.$\downarrow$& C.$\downarrow$ & F.$\uparrow$ & C.$\downarrow$ & F.$\uparrow$ \\
     \midrule
     \midrule
V1   & -  & -& - & 0.021 & 0.045 & 0.768 & 1.24 & 0.065 \\
V2   & \checkmark & - & - & 0.004 & 0.009 & 0.99 & 1.25 & 0.066 \\
V3   & \checkmark & \checkmark & - & 0.003 & 0.009 & 0.99 & 0.045 & 0.735 \\
V4   & - & \checkmark & \checkmark & 0.015 & 0.029 & 0.888 & 0.039 & 0.785\\
Full & \checkmark & \checkmark & \checkmark & \textbf{0.003} & \textbf{0.009} & \textbf{0.99} & \textbf{0.033} & \textbf{0.817} \\
\bottomrule \\
\end{tabular}

	}
	\caption{
	\textbf{Ablation Study.} The D. denotes the depth error of the occluded background. The C. and F. mean the Chamfer-$L_1$ and F-score respectively, for the reconstruction of occluded background regions and full complete objects. Details in Section~\ref{section-ablation}.
}
\label{tab:ablation}
\end{table}

We first quantitatively analyze the effectiveness of the proposed regularizations on synthetic scenes by comparing our full method to four variants in Tab.~\ref{tab:ablation}. 
First, adding the smoothness loss~(V2) significantly decreases the depth error of occluded background regions. 
Next, the $\cL_{\text{op}}$ makes the object reconstruction watertight and the object-level metrics are greatly improved~(V3). The $\cL_{\text{rd}}$ further improves the object reconstruction performance~(Full). Results of V4 prove that without $\cL_{\text{bs}}$ to improve the background, the object reconstruction quality also drops. Note that $\cL_{\text {rd}}$ needs to be applied jointly with $\cL_{\text{op}}$ to guarantee its second condition.

\begin{table}[t]
	\centering
	\resizebox{0.99\linewidth}{!}{%
		\begin{tabular}{l|ccc|cc}
\hline
        & \multicolumn{3}{c}{Background}  & \multicolumn{2}{c}{Objects} \\
        \hline
           & D.$\downarrow$  & C.$\downarrow$ & F.$\uparrow$ & C.$\downarrow$ & F.$\uparrow$ \\
           \hline\hline
ObjSDF     & 0.033      & 0.076   & 0.669   & 1.040        & 0.084       \\
ObjSDF*    & 0.021      & 0.044   & 0.772   & 1.030        & 0.085       \\
ObjSDF*-C & 0.021      & 0.044   & 0.772   & 0.134        & 0.681       \\
Ours       & \textbf{0.003}      & \textbf{0.009}   & \textbf{0.990 }  & \textbf{0.033 }       & \textbf{0.817 }     \\
\hline
\end{tabular}

	}
    \vspace{1em}
	\caption{
	\textbf{Synthetic Scenes Reconstruction Quantitative Results.} The D. C. and F. have the same meanings as in Tab.~\ref{tab:ablation}.
}
\label{tab:synthetic}
\end{table}

A qualitative case is shown in Fig.~\ref{fig:ablation} for better illustration.
In the first row, the occluded part of the background can have many artifacts because of occlusion. 
With the smoothness regularization, the reconstructed backgrounds are much cleaner. Since the $\cL_{\text{bs}}$ only applies to the occluded region, the other parts won't be affected. In the second row, the sofa is pushed against the wall and reconstruction from \cite{wu2022object} contains only visible open surface. The $\cL_{\text{op}}$ first regularizes the SDF field to be a watertight mesh. The $\cL_{\text{rd}}$ further mitigates the flaws at the back by regularizing the object to be confined to the background.

\subsection{Reconstruction in Synthetic Scenes}
\label{exp:synthetic-reconstruction}

Since the synthetic scenes are rendered with objects that have accurate object-level 3D ground truth geometry, we compare our methods with previous object-compositional reconstruction method~\cite{wu2022object} on the $5$ generated scenes and report the quantitative results in Tab.~\ref{tab:synthetic}.

In Tab.~\ref{tab:synthetic} we provide detailed metrics on object-level reconstruction. A qualitative comparison is shown in Fig.~\ref{fig:synthetic}. The computation procedure of all the metrics is available in the supplementary. 
With monocular cues, ObjSDF* can achieve better performance than the original ObjSDF. However, their performance on object reconstruction is far from satisfactory, since they can only accurately obtain the visible surface with large parts of irrelevant structures in the indoor scenes as pointed out in the analysis before. 
Though ObjSDF*-C can eliminate most of the outliers and get improved performance, it can only reconstruct visible regions as an open surface as shown in Fig.~\ref{fig:synthetic}.
On the contrary, our regularizations help to smoothen the background and recover the object as a watertight mesh, which promotes performance by a large margin and leads to broader downstream applications. As shown in Fig.~\ref{fig:synthetic}, RICO can reconstruct the unobservable~(back view as in the caption) regions of the objects where the previous methods fail.

\begin{table}[t]
	\centering
	\resizebox{0.95\linewidth}{!}{%
		\begin{tabular}{lccc|c}
\toprule
{} & ObjSDF & ObjSDF* & Ours & MonoSDF\\
\midrule
\midrule
Chamfer-$L_1$~$\downarrow$ & 0.170 & 0.092 & \textbf{0.088} & 0.090  \\
F-Score~$\uparrow$   & 0.357 & 0.567& \textbf{0.624} & 0.610\\
\bottomrule\\
\end{tabular}

	}
	\caption{
	\textbf{ScanNet Reconstruction Quantitative Results.}
}
\label{tab:scannet}
\end{table}

\subsection{Reconstruction in Real-world Scenes}
\label{exp-scannet}

We conduct experiments on 7 scenes of ScanNet~\cite{dai2017scannet} to show the effectiveness of our method on real-world data. Since ScanNet only provides ground truth for the visible surface, here we follow the protocol in \cite{guo2022neural} and report the reconstruction performance of the entire scene in Tab.~\ref{tab:scannet}. By utilizing the rendering formulation in \cite{wang2021neus} which explicitly models the angle between surface normal and ray, our method can achieve slightly better performance compared to ObjSDF* and \cite{yu2022monosdf} on visible surfaces.

In Tab.~\ref{tab:psnr-miou} we evaluate the 2D rendering and segmentation performance and provide the PSNR and mIoU results on two real-world datasets. As our method constrains mainly the unobservable, the PSNR and mIoU metrics are not with significant difference.

\begin{table}[t]
	\centering
	\resizebox{0.95\linewidth}{!}{%
		\begin{tabular}{ll|c|c|c}
\toprule
                         &      & ObjSDF & ObjSDF* & Ours  \\
                         \midrule
                         \midrule
\multirow{2}{*}{PSNR$\uparrow$} & ToyDesk & 21.10  & 21.21   & \textbf{21.33} \\
                         & ScanNet & 22.78  & 23.15   & \textbf{23.30}  \\ 
                         \midrule
\multirow{2}{*}{mIoU$\uparrow$} & ToyDesk & 0.88   & \textbf{0.89}    & \textbf{0.89} \\
                         & ScanNet & 0.89   & 0.91    & \textbf{0.92} \\
                         \bottomrule\\
\end{tabular}
	}
	\caption{
	\textbf{PSNR and mIoU Results} of different methods on scenes from ScanNet and ToyDesk datasets.
}
\label{tab:psnr-miou}
\end{table}

In Fig.~\ref{fig:scannet} we also provide the comparisons over two scenes. Note that the ScanNet images are blurry and noisy, and the masks are sometimes inaccurate~(\eg, legs of chairs missing), leading to less accurate reconstructions. However, it can still be seen that our method can successfully regularize the unobservable regions to get the watertight mesh, while the baselines always result in open surface. For example, the piano in the first row and the sofas in the second row can only be reconstructed on visible surfaces. 

\begin{figure}[t]
\begin{center}
\includegraphics[width=0.95\linewidth]{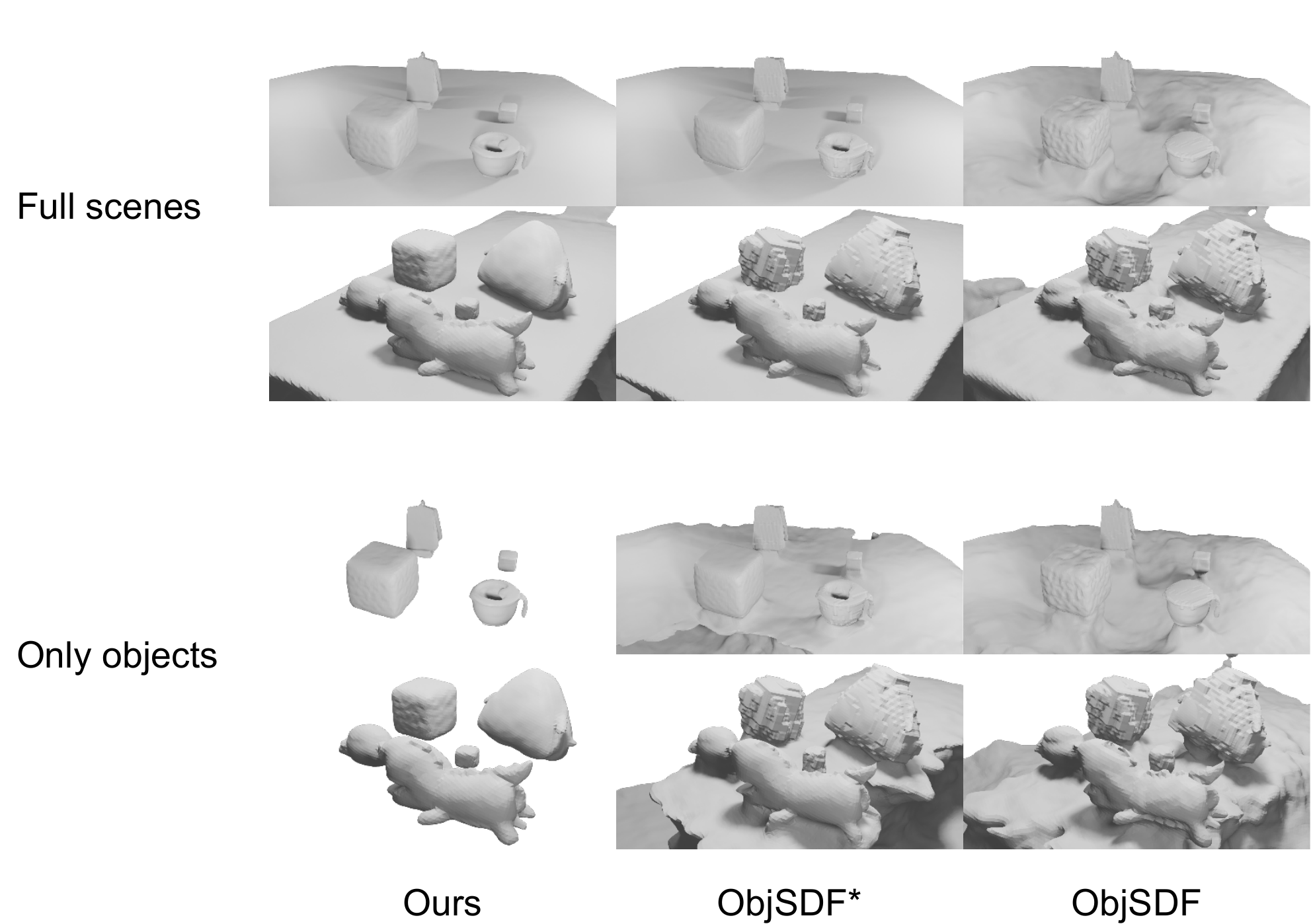}
\end{center}
   \caption{\textbf{Qualitative Results on ToyDesk.} Here we show the reconstruction results both with and without the background for better visualization.}
\label{fig:toydesk}
\end{figure}

And in Fig.~\ref{fig:toydesk} we provide the qualitative results of our method and two baselines on two ToyDesk scenes. It can be seen that the baseline methods generate non-watertight surface with undesired artifacts~(using table plane to cut the mesh will result in holes on the cut face), while our method can directly get clean results.

\subsection{Object Manipulation}

\begin{figure}[t]
\begin{center}
\includegraphics[width=\linewidth]{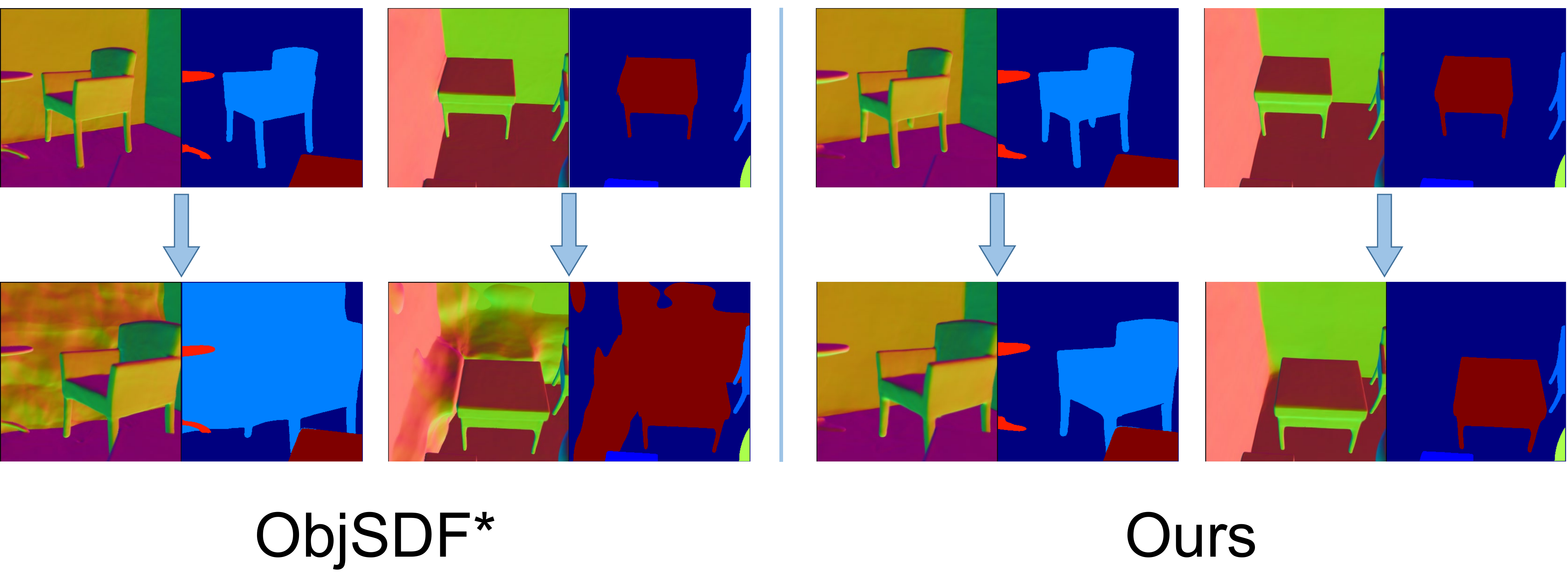}
\end{center}
   \caption{\textbf{Object Manipulation Cases.} Here we show the volume rendered normal maps and segmentation masks in two scenes, before~(top row) and after~(bottom row) the manipulation.}
\label{fig:manipulation}
\end{figure}

As aforementioned, the reconstruction results from ObjSDF* are sub-optimal for downstream applications like object manipulation. On the contrary, since RICO can get a watertight mesh, it can be easily used for such applications.
In Fig.~\ref{fig:manipulation} we show the volume rendered normal maps and segmentation masks before and after moving an object in the scene. An illustration of how to manipulate one object in our framework and conduct volume rendering accordingly is applicable in the supplementary. As illustrated, after manipulation, the rendered segmentation and normal map are still clean for RICO. In contrast, the results of ObjSDF* are messy because their reconstruction is connected with artifacts that were originally outside of the scenes.

\section{Conclusion}
\label{conclusion}
We have presented RICO, a novel approach for compositional reconstruction in indoor scenes. 
Our key motivation is to regularize the unobservable regions for the objects with partial observations in indoor scenes. 
We exploit the geometry smoothness for the occluded background, and then adopt the improved background as the prior to regularize the objects' geometry. Our experimental results prove that our method achieves compositional reconstruction with fewer artifacts and watertight object geometry, which further facilitates applicable applications like object manipulation.

\boldparagraph{Future work} 
Currently, each object's pose is baked with its geometry. An interesting future direction would be represent each object in scene as the combination of its own canonical coordinates and the pose, \ie the rotated 3D bounding box in the world coordinate. With these disentangled representations, we can cast more regularizations from the shape and box perspectives, and the disentanglement also allows more flexible downstream applications.
Another interesting direction would be developing more semantic-based prior knowledge for systemic regularization. Now the geometry motivated regularizations are lack of the understanding of the actual shape of each object. With the general knowledge emerged in recent large vision-language model, like StableDiffusion, it's interesting to distill its knowledge in our compositional reconstruction task. The learnt semantic knowledge can provide a holistic understanding of each object and provide a more comprehensive regularization.

\boldparagraph{Acknowledgements} We thank Kechun Xu and Huaijin Pi for detailed feedback on the paper. We thank all authors and reviewers for the contributions. This work is in part supported by a Grant from the National Natural Science Foundation of China~(No.U21A20484), the NSFC under grant U21B2004, 62202418, and the Zhejiang University Education Foundation Qizhen Scholar Foundation.

\clearpage

{\small
\bibliographystyle{ieee_fullname}
\bibliography{bibliography_long, egbib}
}

\clearpage

\appendix

\section{Network Architecture}
\label{supp-architecture}

\begin{figure}[ht]
\begin{center}
\includegraphics[width=\linewidth]{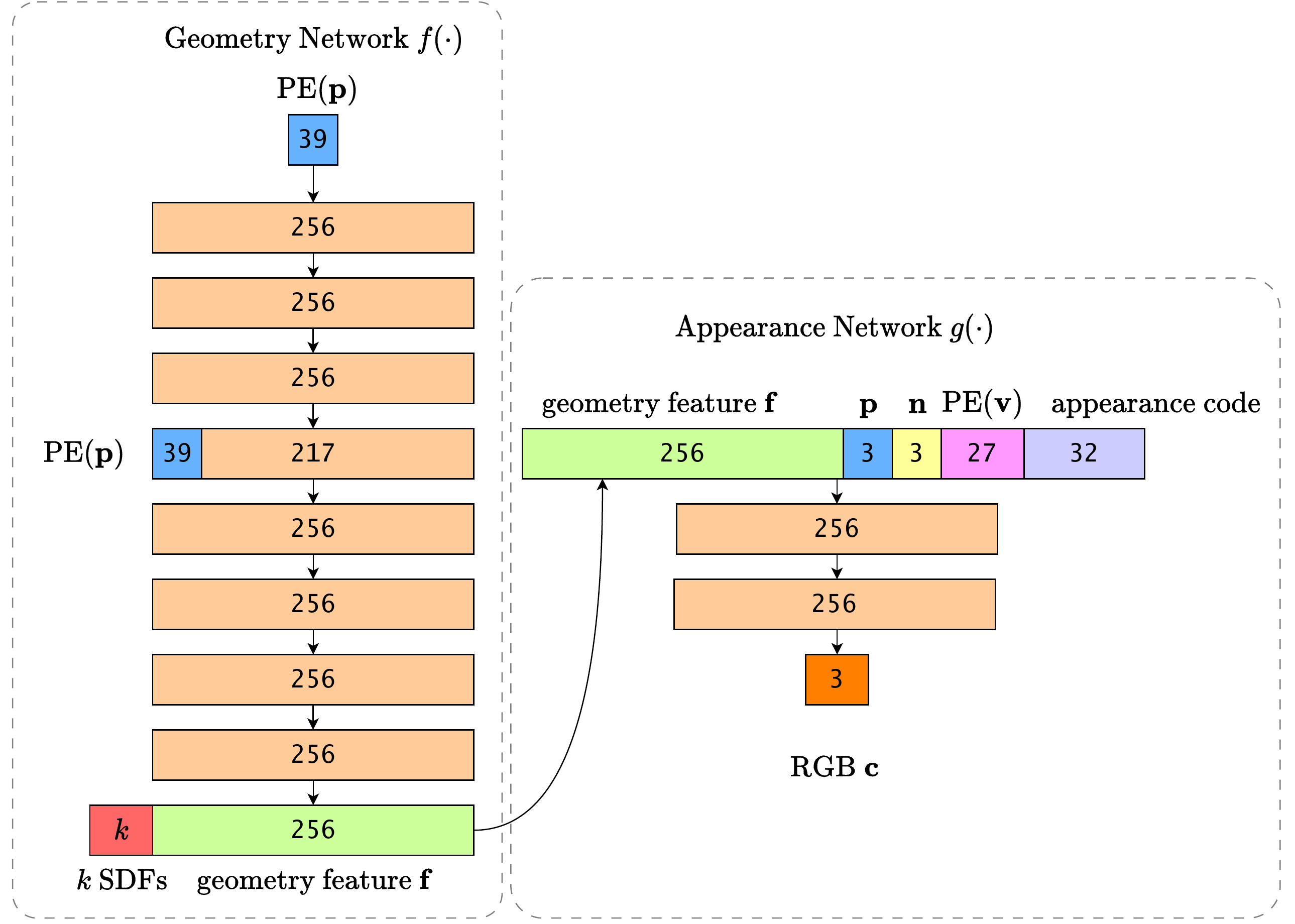}
\end{center}
   \caption{\textbf{Network Architecture.} The geometry network corresponds to the SDF function $f$ and the appearance network denotes the appearance $g$ in the main paper. }
\label{fig:network}
\end{figure}

We adopt the similar network structure as in \cite{wu2022object} and add the appearance code utilized in \cite{yu2022monosdf} for all the baseline models and ours. Fig.~\ref{fig:network} shows our detailed network architecture.

The geometry network is a 8-layer MLP with hidden dimension 256, and one skip connection at the fourth layer. The input of this network is the point coordinate mapped by a fixed positional encoding~\cite{mildenhall2021nerf}. The output consists of a 256 dimensional geometry feature and $k$ SDF values for each object. These SDF values can be transformed to semantic logits using the function proposed in \cite{wu2022object}.

For the input of the appearance network, we adopt the design in \cite{yu2022monosdf}. We optimize a per-frame appearance code during the training and use this per-frame code to model the varying light and blurry condition in each image for better reconstruction. The appearance code is concatenated with the viewing direction~(also mapped by positional encoding), the normal of the scene SDF, the geometry feature and the point coordinate. The appearance network consists of two layers with hidden dimension 256, and outputs a 3-channel RGB color.

We use Softplus activation for the geometry network and use ReLU
activation for the appearance network, the RGB color is obtained after passing the network's output through the Sigmoid activation.

\section{Loss Functions Details}

We elaborate all the losses and the weight choices used in the optimization in this section.

\boldparagraph{RGB Reconstruction Loss}
To learn the surface from images input, we need to minimize the difference between ground-truth pixel color and the rendered color. We follow the previous works~\cite{yu2022monosdf,wu2022object} here for the RGB reconstruction loss:

\begin{equation}
    \cL_{\text{RGB}} = \sum_{\br}||\hat{\bC}(\br) - \bC(\br)||_1.
\end{equation}
Here $\hat{\bC}(\br)$ is the rendered color from volume rendering and $\bC(\br)$ denotes the ground truth.

\boldparagraph{Depth Consistency Loss}
Monocular depth and normal cues~\cite{yu2022monosdf} can greatly benefit indoor scene reconstruction. For the depth consistency, we minimize the difference between rendered depth $\hat{D}(\br)$ and the depth estimation $\bar{D}(\br)$ from the Omnidata~\cite{eftekhar2021omnidata} model:

\begin{equation}
    \cL_{\text{D}} =  \sum_{\br}|| (w\hat{D}(\br) + q) - \bar{D}(\br) ||^2,
\end{equation}
where $w$ and $q$ are the scale and shift values to match the different scales. We solve $w$ and $q$ with a least-squares criterion, which has the closed-form solution. Please refer to the supplementary of \cite{yu2022monosdf} for a detailed computation process.

\boldparagraph{Normal Consistency Loss}
Similar to the depth consistency loss, we also use the normal cues $\bar{N}$ from Omnidata model to supervise the rendered normal. Specifically, the normal consistency loss consists of L1 and the angular losses:
\begin{equation}
    \cL_{\text{N}} = \sum_{\br} ||\hat{N}(\br) - \bar{N}(\br)||_1 + ||1-  \hat{N}(\br)^{\text{T}}\bar{N}(\br)||_1.
\end{equation}
Here the volume-rendered normal and normal estimation will be transformed into the same coordinate system by the camera pose.

\boldparagraph{Semantic Loss}
We minimize the semantic loss between volume-rendered semantic logits of each pixel and the ground-truth pixel semantic class. Here the semantic objective is implemented as a cross-entropy loss:
\begin{equation}
    \cL_{\text{S}} = \sum_{\br}\sum_{j=1}^k-\hat{h}_j(\br)\log h_j(\br).
\end{equation}
The $\hat{h}_j(\br)$ is the ground-truth semantic probability for $j$-th object, which is $1$ or $0$.

\boldparagraph{Eikonal Loss}
Following common practice, we also add an Eikonal term on the sampled points to regularize SDF values in 3D space:
\begin{equation}
    \cL_{\text{E}} = \sum_{i}^{n}(|| \nabla \min_{1\leq j\leq k}~s_j(\bp_i)||_2 - 1)
\end{equation}
Here the eikonal loss is applied to the gradient of the scene SDF, which is the minimum of all the SDFs.


\section{Evaluation Metrics}

To evaluate the reconstruction performance, we use the Chamfer Distance and F-score with a threshold of 5cm in this paper. In detail, Chamfer Distance comes from \textit{Accuracy} and \textit{Completeness}, and F-score is derived from \textit{Precision} and \textit{Recall}. For point clouds $P$ and $P^*$ sampled from the predicted and the ground-truth mesh, we show the detailed computation procedure here:

\small
\begin{equation}
    \begin{aligned}
        \texttt{Accuracy} &= \mathop{\texttt{mean}}_{\bp \in P}\left( \min_{\bp^*\in P*} ||\bp-\bp^*||_1 \right), \\
        \texttt{Completeness} &= \mathop{\texttt{mean}}_{\bp^* \in P^*}\left( \min_{\bp\in P} ||\bp-\bp^*||_1 \right), \\
        \texttt{Chamfer-}L_1 &= \frac{\texttt{Accuracy} + \texttt{Completeness}}{2}.
    \end{aligned}
\end{equation}

\begin{equation}
    \begin{aligned}
        \texttt{Precision} &= \mathop{\text{mean}}_{\bp \in P}\left( \min_{\bp^*\in P*} ||\bp-\bp^*||_1 < 0.05 \right), \\
        \texttt{Recall} &= \mathop{\text{mean}}_{\bp^* \in P^*}\left( \min_{\bp\in P} ||\bp-\bp^*||_1 < 0.05 \right), \\
        \texttt{F-score} &= \frac{2\times \texttt{Precision}\times \texttt{Recall}}{\texttt{Precision}+ \texttt{Recall}}.
    \end{aligned}
\end{equation}
\normalsize

\section{Synthetic Dataset Construction}

In order to quantitatively evaluate the object-level reconstruction performance in the object-compositional indoor scenes, we create a synthetic dataset with object ground-truth geometry. In this part we elaborate on how to construct the Synthetic Dataset used in this paper. Despite that the dataset is not a major contribution of this paper, we will release it for future comparisons.

We use Blender~\cite{blender} and an add-on BlenderNeRF~\cite{Raafat_BlenderNeRF_2023} to construct the scenes~(assign different object locations, lighting conditions and camera trajectories) and render the RGB images together with the camera poses. The Blender's camera coordinate system is different from the coordinate system in ScanNet, which requires an extra $180^\circ$ rotation along with the x-axis on the recorded extrinsic matrix. 

To render semantic masks, we switch each object's surface texture to a certain value and render again with the identical camera trajectories. We create $5$ scenes and three of them contain $5$ objects while other two contain $10$ objects~(background not included). For each scene we render $200$ images and use the Omnidata~\cite{eftekhar2021omnidata} model to obtain the corresponding monocular depth and normal cues.

\section{Additional Ablation Experiments}

\subsection{Parameter Ablation Study}
In this part we provide the ablation experiment for the $\epsilon$ in proposed object point-SDF loss $\cL_{\text{op}}$. Particularly, $\epsilon$ is a non-negative number as a threshold, that the objects' SDFs outside of the background should be larger than this value. We provide an ablation study on different $\epsilon$ values on synthetic scenes in Tab.~\ref{tab:epsilon}.

\begin{table}[htbp]
	\centering
	\resizebox{0.95\linewidth}{!}{%
		\begin{tabular}{lcccc}
\toprule
{} & $\epsilon = 0$ & $\epsilon = 0.05$ & $\epsilon = 0.1$ &$\epsilon = 0.2$\\
\midrule
\midrule
Chamfer-$L_1$~$\downarrow$ & 0.187 & \textbf{0.033} & 0.034 & 0.036  \\
F-Score~$\uparrow$   & 0.755 & \textbf{0.817} & 0.812 & 0.793 \\
\bottomrule\\
\end{tabular}

	}
	\caption{
	\textbf{Ablation Study on $\epsilon$.} Metrics are evaluated and averaged on all the objects of all the synthetic scenes.
}
\label{tab:epsilon}
\end{table}

Intuitively, $\epsilon$ should be larger than 0 because the points behind the background are outside of each object's surface, \ie the object SDFs of these points should be positive. Empirically we find setting $\epsilon=0.05$ yields slightly better performance than $0.1$ and $0.2$. When setting $\epsilon=0$, the object reconstruction performance drops significantly. We found the reason is that, when $\epsilon$ is $0$, the SDFs of the sampled points can not all be effectively optimized to positive, yielding some negative SDF vaules, which results in the flaws in the empty space.

\subsection{Backbone Ablation Study}

When utilizing SDF as the surface geometry representation, there are typically two choices to combine SDF and volume rendering as proposed in\cite{yariv2021volume,wang2021neus}. In the main paper we adopt the scheme proposed in NeuS~\cite{wang2021neus} for RICO. We provide the comparison of reconstruction results on ScanNet~(evaluated on whole scene) and Synthetic scenes~(evaluated on each object), and report the results in Tab.~\ref{tab:backbone}. Since \cite{wang2021neus} explicitly models the angle difference of ray direction and the surface normal, it can have slightly better performance by better reconstructing the visible surface.

\begin{table}[htbp]
    \centering
    \resizebox{0.95\linewidth}{!}{%
		\begin{tabular}{lcccc}
\toprule
            & \multicolumn{2}{c}{ScanNet} & \multicolumn{2}{c}{Synthetic Object} \\
            & Chamfer-$L_1$ $\downarrow$  & F-score $\uparrow$  & Chamfer-$L_1$  $\downarrow$      & F-score   $\uparrow$    \\
\midrule
\midrule
RICO-VolSDF & 0.090           & 0.592     & 0.042                & 0.751         \\
RICO        & 0.088           & 0.624     & 0.033                & 0.817 \\
\bottomrule \\
\end{tabular}

	}
    \caption{\textbf{Ablation Study on Backbone}. We show the reconstruction comparison of our methods using the volume rendering scheme proposed in \cite{yariv2021volume}~(RICO-VolSDF) and \cite{wang2021neus}~(RICO, which represents the method proposed in main paper).}
    \label{tab:backbone}
\end{table}

\section{Construct the ObjSDF*-C Baseline}

As stated in the main paper, we construct an improved baseline over ObjSDF*, named ObjSDF*-C, to provide better visualization and quantitative results. The main procedure is to use the reconstructed background surface to eliminate the parts of object reconstruction that are outside of the background range. The construction procedure is just a post-process method on the object meshes and do not change the original nature of the ObjSDF* that only the visible surfaces are reconstructed.

For Synthetic scenes, since we set the background as a cubic room with a range of $[-2m, 2m]$ in three dimensions, \ie the background is an axis-aligned box, we can directly use this range to segment the object reconstruction meshes. For ScanNet scenes, we use the ground-truth scene meshes to get a coarse range and manually finetune the range of each scene~(ObjSDF*-C on ScanNet is only used for visualization, not for quantitative evaluation), then segment the object meshes based on the finetuned range.

\section{Object Manipulation Implementation}
\label{manipulation}

\begin{figure}[ht]
\begin{center}
\includegraphics[width=\linewidth]{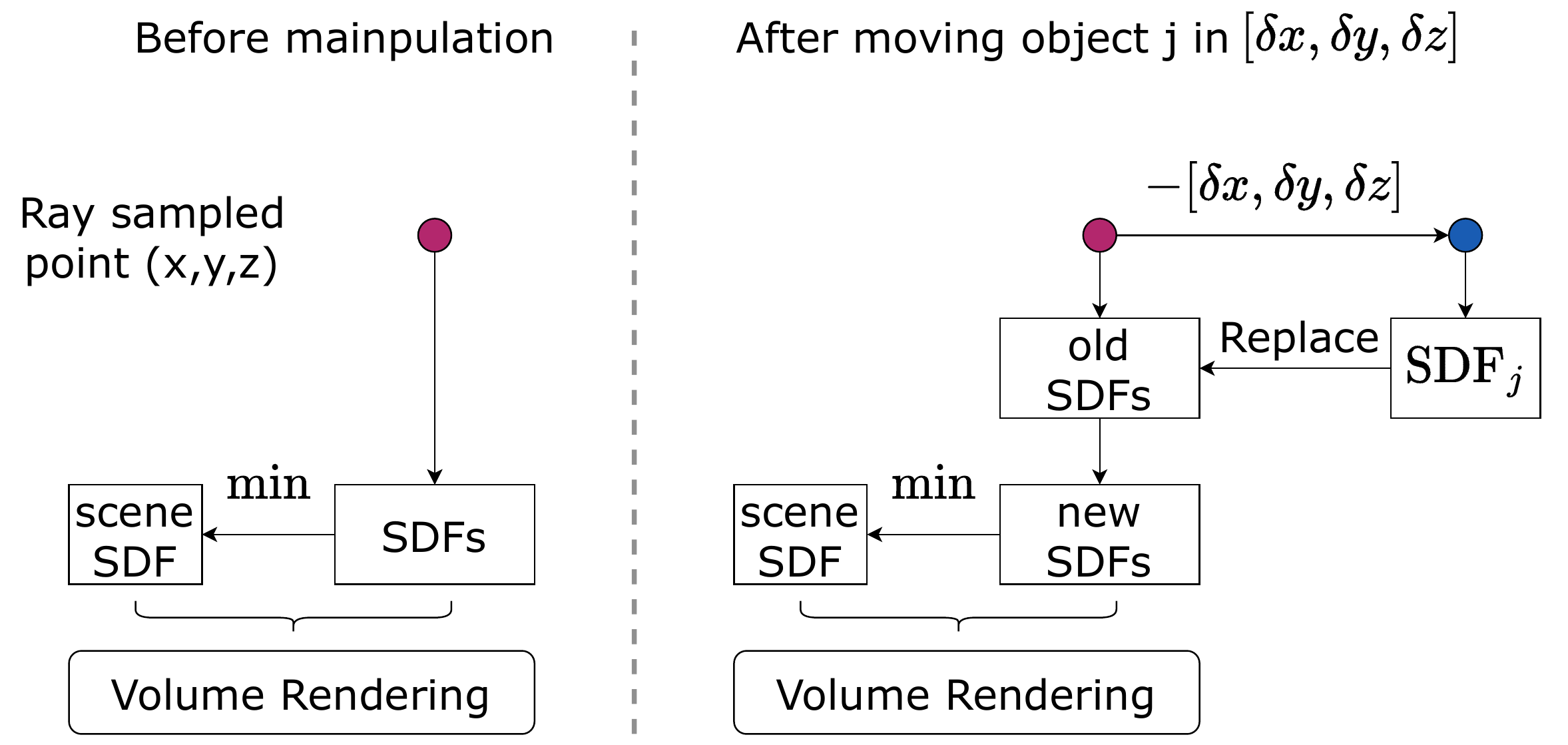}
\end{center}
   \caption{\textbf{Object Manipulation Implementation.} In this figure we show when moving the object $j$ with $\left[\delta x, \delta y, \delta z\right]$ in each direction, how to implement volume rendering in the current network.}
\label{fig:manipulation-implementation}
\end{figure}

To manipulate the reconstructed objects, a straightforward way is to directly manipulate the meshes. In the main paper we show the volume rendered normal maps and semantic masks before and after manipulation. In Fig.~\ref{fig:manipulation-implementation} we show how to implement the volume rendering in current framework. The core is to query the SDF value of manipulated object at the destined point, and combine it with other objects' SDF values. 
Notably, the color is decided by not only the coordinate but also the geometry feature~(illustrated in Fig.~\ref{fig:network}).
However, now the original point for other object SDFs and the manipulated point for the desired object SDF will result in two geometry feature vectors. 
In contrast to use minimum value to get the scene SDF from all SDFs, it's hard to decide how to fuse these two geometry features together in current framework. Now we only show the volume rendering results that are decided by SDF values, \ie geometry, like the normal map and semantic mask.

\section{More Discussions}

\boldparagraph{Smooth surface} 
It can be seen from the figures in the paper that surface from RICO is smoother in a way. The smooth effect actually comes from different rendering schemes proposed in VolSDF~\cite{yariv2021volume} and NeuS~\cite{wang2021neus}. Empirically we notice that comparing to RICO-VolSDF, RICO-NeuS (default) has better quantitative results but can also be somehow oversmooth. It is an interesting future direction to investigate the effects of  different reconstruction backbones in learning the compositional geometry.

\begin{figure}[t]
\begin{center}
\includegraphics[width=0.91\linewidth]{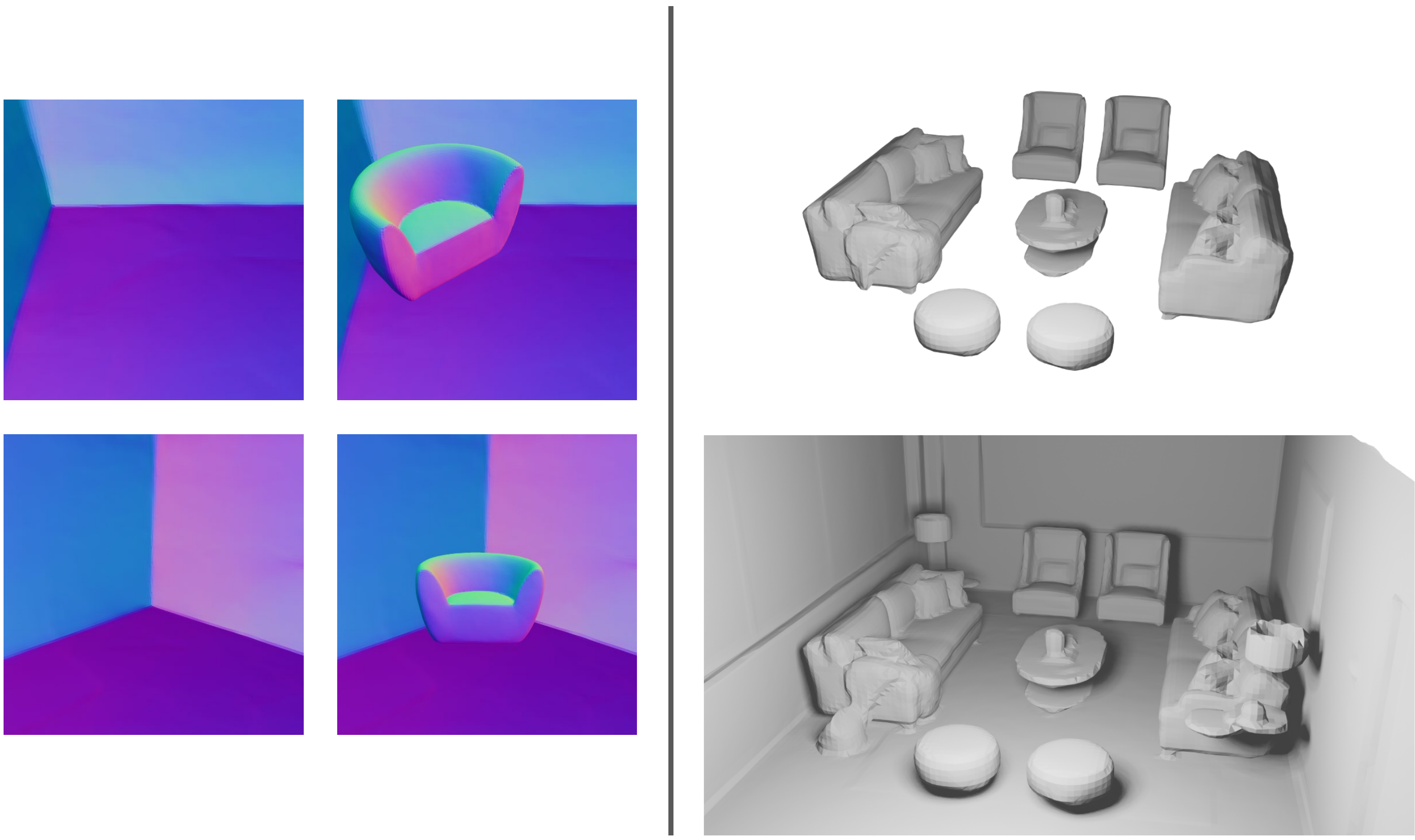}
\end{center}
\caption{\textbf{Visualizations of: } (left)~normal map of occluded corner in background, (right)~reconstruction results on Replica.}
\vspace{-1.5em}
\label{fig:vis_bgocc_replica}
\end{figure}

\boldparagraph{Background smoothness regularization}
Fig.~\ref{fig:synthetic} of our paper presents sharp background corners with occlusions. Here we provide a better visualization in Fig.~\ref{fig:vis_bgocc_replica} (left). 
There are mainly two scenarios for smoothness loss on occluded sharp geometry: 1)~The sharp geometry is observable in some images. The reconstruction loss in these views will be dominant in optimization because the smoothness loss is of small weight and computed only once in several iterations, thus the sharp geometry can be reconstructed correctly; 2)~The sharp geometry is completed occluded in all the views as shown in Fig.~\ref{fig:vis_bgocc_replica}. By regularizing the depth and normal, we observe that the visible regions of wall and ground are smoothly extended, yielding a corner that is not perfectly perpendicular but without artifacts.

\section{Qualitative Results on Individual Scenes}

\begin{figure*}[ht]
    \centering
    \includegraphics[width=\linewidth]{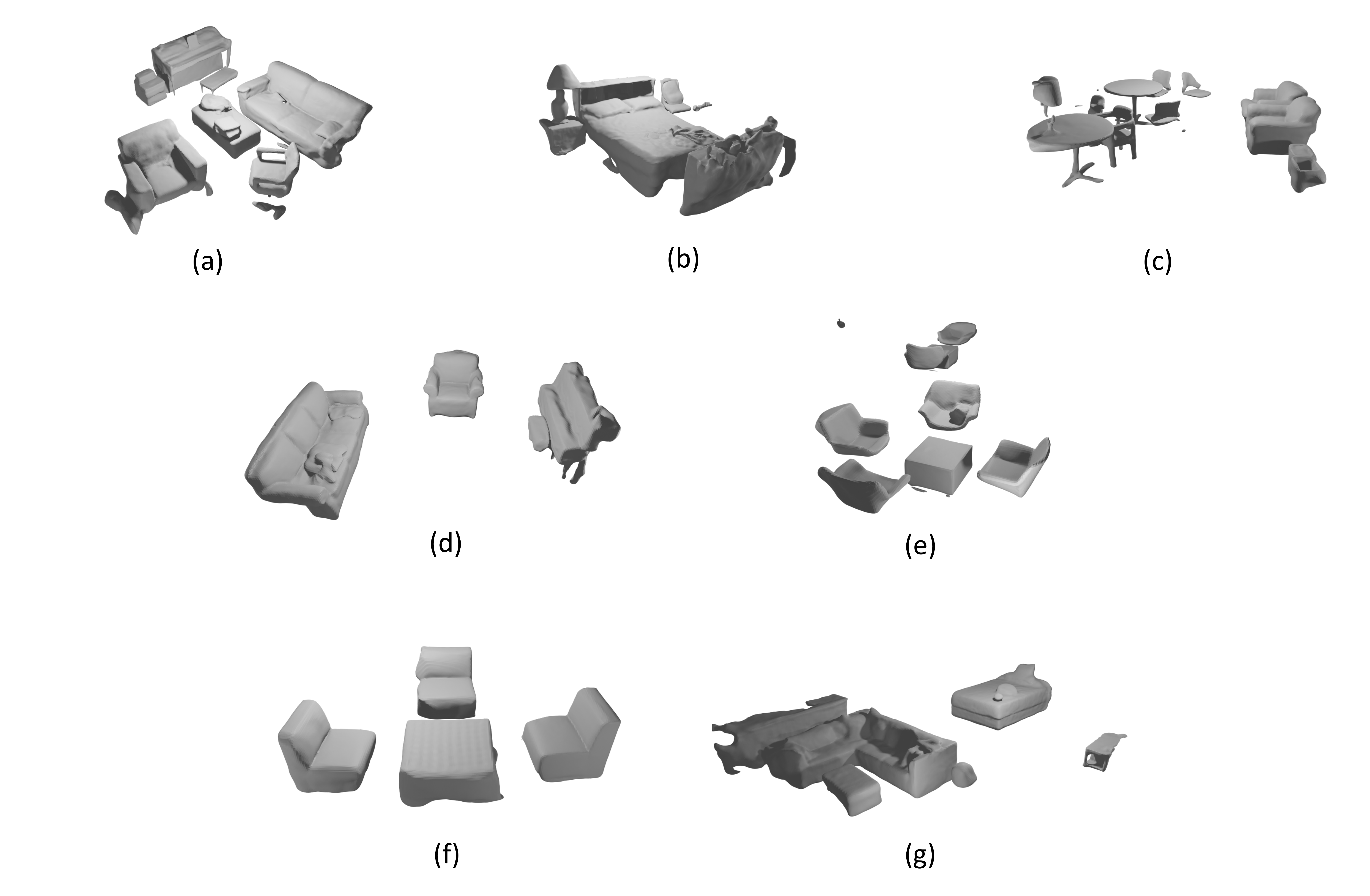}
    \caption{\textbf{Qualitative Visulization on ScanNet.} We show the object-compositional reconstruction results from RICO on seven ScanNet~\cite{dai2017scannet} scenes.}
    \label{fig:vis_scannet}
\end{figure*}

\begin{figure*}[htbp]
    \centering
    \includegraphics[width=0.85\linewidth]{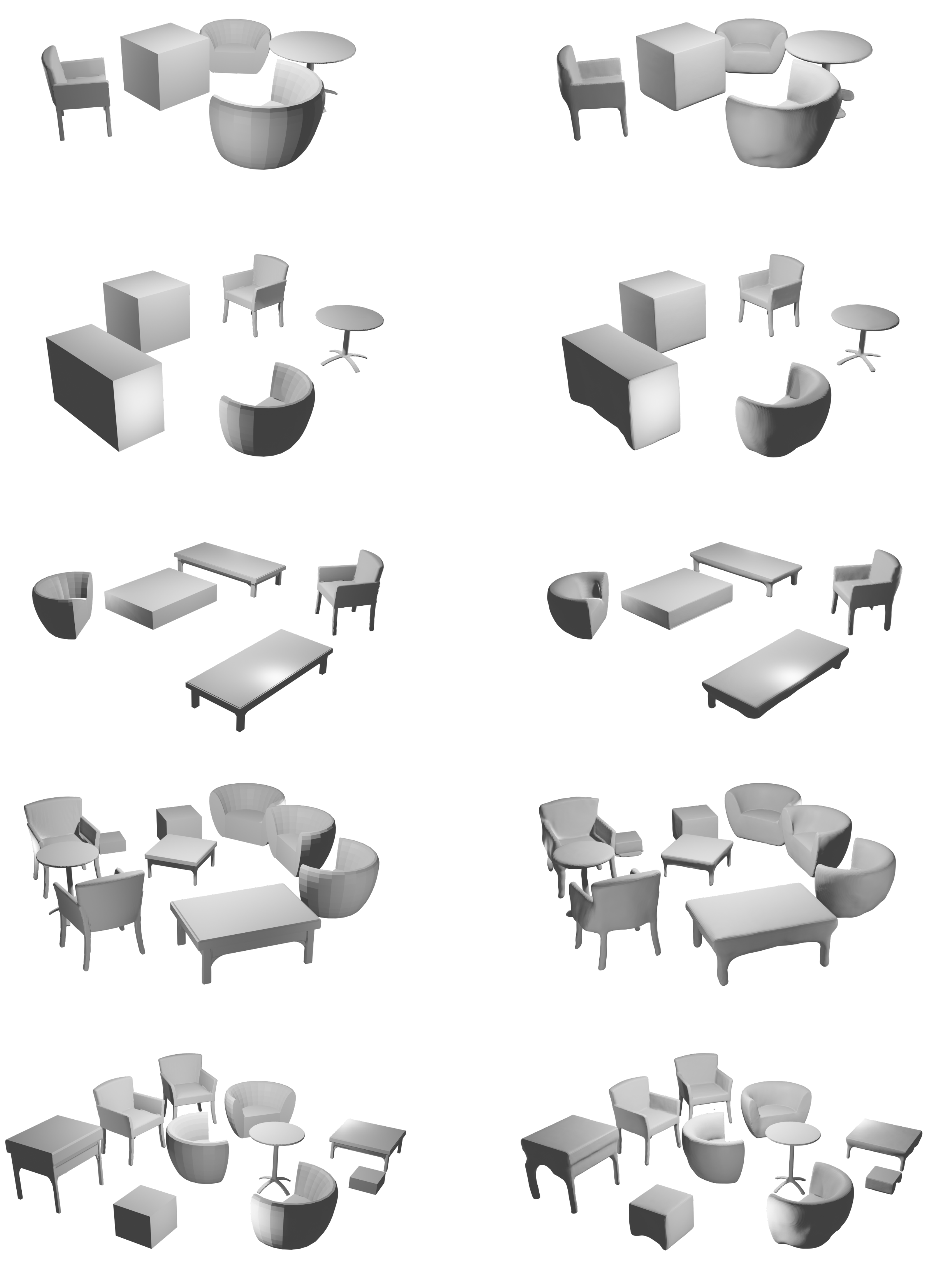}
    \caption{\textbf{Qualitative Visualization on Synthetic Scenes.} In the left column we show the ground-truth object geometry of the five synthetic scenes, in the right column we provide the qualitative object-compositional reconstruction results of our proposed RICO.}
    \label{fig:vis_synthetic}
\end{figure*}

In Fig.~\ref{fig:vis_scannet} and Fig.~\ref{fig:vis_synthetic}, we provide the RICO's object-compositional reconstruction on ScanNet scenes and Synthetic scenes~(with the object ground-truth) respectively. Here we also provide a visualization of RICO on one of the Replica scenes in Fig.~\ref{fig:vis_bgocc_replica}~(right).

\section{Limitations} 
In this work, we assume the indoor scene as convex room, that the ray shot inside of the room penetrates the background surface once. 
However when processing the more complex indoor scenes where one ray can go through multiple rooms, our object regularizations may require extra conditions to decide which points to be applied to.
Additionally, the object-scene relation prior regularized the completeness from only the geometry perspective. The framework can be extended to utilizing more complex category-level prior for better reconstruction.

\end{document}